\documentclass[10pt,twocolumn,letterpaper]{article}

\usepackage{wacv}
\usepackage{times}
\usepackage{epsfig}
\usepackage{graphicx}
\usepackage{amsmath}
\usepackage{amssymb}

\usepackage[utf8]{inputenc} 
\usepackage[T1]{fontenc}    
\usepackage{url}            
\usepackage{booktabs}       
\usepackage{amsfonts}       
\usepackage{nicefrac}       
\usepackage{microtype}      
\usepackage{xcolor}         
\usepackage{stmaryrd}
\usepackage{pifont}

\newcommand{\xmark}{\ding{55}}
\usepackage{multirow}
\usepackage{wrapfig}
\usepackage{booktabs}
\usepackage{subcaption}
\usepackage[accsupp]{axessibility}
\usepackage{longtable}
\usepackage{array}

\wacvfinalcopy

\usepackage[pagebackref=true,breaklinks=true,colorlinks,bookmarks=false]{hyperref}

\begin{document}

\title{Multi-Task Classification of Sewer Pipe Defects and Properties using a Cross-Task Graph Neural Network Decoder}

\author{Joakim Bruslund Haurum$^1$ \qquad Meysam Madadi$^{2}$ \qquad Sergio Escalera$^{1,2,3}$ \qquad Thomas B. Moeslund$^1$\\[0.25em]
{\normalsize $^1$ Visual Analysis and Perception (VAP) Laboratory, Aalborg University, Denmark}\\
{\normalsize$^2$ Computer Vision Center, Autonomous University of Barcelona, Spain}\\ 
{\normalsize$^3$ Dept. of Mathematics and Informatics, Universitat de Barcelona, Spain}\\
{\tt\small joha@create.aau.dk, mmadadi@cvc.uab.es, sergio@maia.ub.es, tbm@create.aau.dk}
}

\maketitle

\thispagestyle{empty}

\begin{abstract}
The sewerage infrastructure is one of the most important and expensive infrastructures in modern society. In order to efficiently manage the sewerage infrastructure, automated sewer inspection has to be utilized. However, while sewer defect classification has been investigated for decades, little attention has been given to classifying sewer pipe properties such as water level, pipe material, and pipe shape, which are needed to evaluate the level of sewer pipe deterioration. 

In this work we classify sewer pipe defects and properties concurrently and present a novel decoder-focused multi-task classification architecture Cross-Task Graph Neural Network (CT-GNN), which refines the disjointed per-task predictions using cross-task information. The CT-GNN architecture extends the traditional disjointed task-heads decoder, by utilizing a cross-task graph and unique class node embeddings. The cross-task graph can either be determined a priori based on the conditional probability between the task classes or determined dynamically using self-attention. CT-GNN can be added to any backbone and trained end-to-end at a small increase in the parameter count. We achieve state-of-the-art performance on all four classification tasks in the Sewer-ML dataset, improving defect classification and water level classification by 5.3 and 8.0 percentage points, respectively. We also outperform the single task methods as well as other multi-task classification approaches while introducing 50 times fewer parameters than previous model-focused approaches. The code and models are available at the project page \url{http://vap.aau.dk/ctgnn}.
\end{abstract}

  \begin{figure}[!t]
  \begin{subfigure}{0.49\linewidth}
      \centering
      \includegraphics[width=\linewidth]{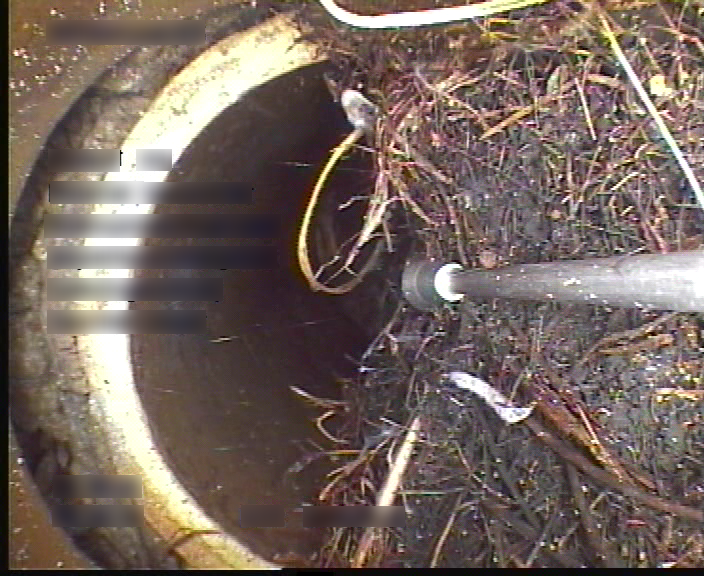}
    \qquad
    \resizebox{\linewidth}{!}{%
    \begin{tabular}[b]{cccc}\toprule
      Task & Ground Truth  & R50-MTL & CT-GNN \\ \cmidrule{1-4}
       Defect & FS, RO & {\color{red}FS}  & FS, RO\\
       Water & [0\%,5\%) & [0\%,5\%) & [0\%,5\%) \\
       Shape & Circular & Circular & Circular \\
       Material & VC & VC & VC\\ \bottomrule
    \end{tabular}%
    }
    \captionlistentry[table]{A table beside a figure}
      \label{fig:FSRO}
  \end{subfigure}
  \begin{subfigure}{0.49\linewidth}
      \centering
      \includegraphics[width=\linewidth]{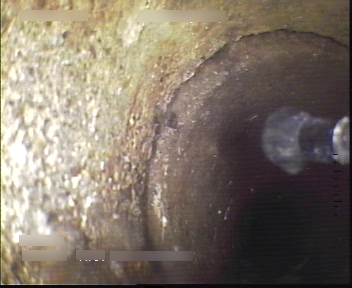}
    \qquad
    \resizebox{\linewidth}{!}{%
    \begin{tabular}[b]{cccc}\toprule
      Task & Ground Truth  & R50-MTL & CT-GNN \\ \cmidrule{1-4}
       Defect & OB, FS & {\color{red}FS}  & OB, FS\\
       Water & [0\%,5\%) & [0\%,5\%) & [0\%,5\%) \\
       Shape & Circular & Circular & Circular \\
       Material & Conc. & {\color{red} VC}  & Conc.\\ \bottomrule
    \end{tabular}%
    }
    \captionlistentry[table]{A table beside a figure}
      \label{fig:OBFS}
  \end{subfigure}
  \caption{Example images from the Sewer-ML dataset \cite{HaurumDataset2021} together with examples showing how the baseline R50-MTL model with no cross-task relationship modeling misses the noticeable roots (RO) and surface damage (OB). Additionally, the R50-MTL model misclassifies the material as vitrified clay (VC) instead of as concrete (Conc.), whereas the proposed CT-GNN model classifies all classes in each task correctly in both examples.}
  \label{fig:examples}
  \end{figure}
\section{Introduction}
The sewerage infrastructure is a key infrastructure of modern society, which needs to be regularly inspected and maintained in order to ensure its functionality \cite{ASCEReportCard2017}. These inspections require professional sewer inspectors who are capable of documenting and differentiating the fine-grained sewer defects, but also the properties of the sewer pipe such as the water level, pipe shape and pipe material, see Figure~\ref{fig:examples}. All of this information can be combined to compute a single deterioration score for each sewer pipe \cite{EUStandard} used by water utility companies for asset management. Due to the hidden nature of the sewerage infrastructure sewer inspections are hard and cumbersome to conduct, as the sewer inspectors have to inspect using a remote controlled vehicle with a movable camera. Each inspection can stretch over a long duration of time due to obstacles in the sewers and limited speed of the vehicle. This leads to prolonged duration of looking at a screen, and can potentially result in flawed inspections due to fatigue.

\begin{figure*}[!htp]
    \centering
    \includegraphics[width=0.89\linewidth]{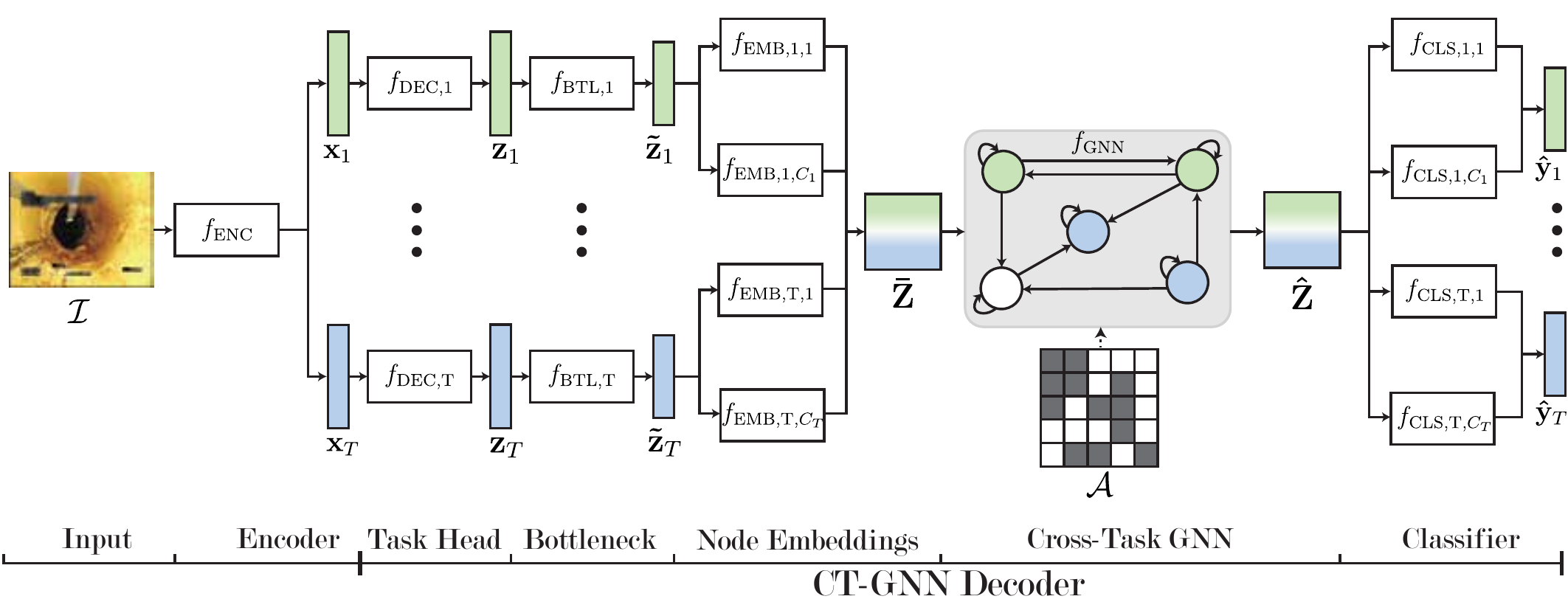}
    \caption{\textbf{CT-GNN Overview.} The proposed CT-GNN decoder and its location within the typical MTC architecture. The initial task features, $\mathbf{z}_t,\ t= 1,2,\dots,T$, from $T$ disjointed task-heads, are refined using our CT-GNN decoder, which incorporates class relationship knowledge, resulting in the final class predictions $\mathbf{\hat{y}}_t,\ t= 1,2,\dots,T$. The CT-GNN is explained in detail in Sections~\ref{sec:ctgnn} \& \ref{sec:adj}.}
    \label{fig:modelOverview}
\end{figure*}
In order to alleviate and assist the sewer inspectors, academia and industry have researched how to automate parts of the inspection process for more than 30 years \cite{HaurumSurvey2020}. However, the majority of work within this field has been focused on the important task of classifying the defects present in the pipes, while omitting the concurrent tasks of determining the water level, pipe material, and pipe shape needed to determine the deterioration score \cite{HaurumSurvey2020}. Furthermore, as the inspections are performed on location it is infeasible to deploy several large models for each task.

Therefore, we investigate how to utilize Multi-Task Learning (MTL), and its sub-field Multi-Task Classification (MTC), to simultaneously classify the sewer pipe defects and properties, by training a single model that is capable of processing multiple tasks during a single forward pass \cite{Vandenhende2021MTL}.

The MTC problem is often defined as learning how to solve several \textbf{unrelated} datasets with a single network \cite{Omniglot2015, VisDeca2017}, whereas the problem of \textbf{related} and \textbf{concurrent} classification tasks, as \eg during sewer inspections, is less well understood \cite{HaurumDataset2021, CelebA2015}. The occurrence of the different task classes follows a hidden intractable joint distribution over all classes from all tasks. While the joint distribution is intractable, the co-occurrence information of the task classes can be inferred from the data, or learned by a model, and subsequently utilized to improve the classification process.

In order to handle the concurrent MTC problem, we propose a novel decoder-focused model, the Cross-Task Graph Neural Network (CT-GNN) Decoder, where the per-task features are refined using a cross-task sharing mechanism, inspired by recent dense vision decoder-focused models \cite{MTI2020, PAD2018, PAP2019, PSD2020}. Specifically, we propose applying a CT-GNN on the initial task feature representations utilizing cross-task class relationships to refine the predictions.

We find that classification of all tasks can be improved by incorporating these cross-task class relationships into the decoder, by either utilizing the a priori known co-occurrence of the different task classes or dynamically estimating it through self-attention. Our proposed method is illustrated in Figure~\ref{fig:modelOverview}. Compared to the previously limited use of graphs in MTC, we do not utilize feature vectors from different images in a batch \cite{AugFeat2020, RelAtt2018} nor do we consider sequential data inputs \cite{MTLSeq2019}. Compared to previous decoder-focused MTC models, we neither estimate the statistical relationship from batches \cite{TaskRelation2019}, nor impose tensor-based constraints \cite{Multilinear2017, TensorFact2017}.

Our contributions are therefore the following:
\begin{itemize}
    \itemsep0em
    \item We present the Cross-Task GNN Decoder, a novel MTC decoder that refines the per-task features through a late cross-task mechanism, trained in an end-to-end manner with only a small parameter count increase.
    \item In order to quantify a priori knowledge of task relationships we construct a cross-task graph adjacency matrix in a data-driven manner.
    \item We achieve State-of-the-Art performance on all four classification tasks in the Sewer-ML dataset \cite{HaurumDataset2021}, demonstrating the importance of utilizing cross-task relationships during automated sewer inspections.
\end{itemize}

The paper is structured as follows. In Section~\ref{sec:related}, we review the related works within the automated sewer inspection as well as MTL and MTC  fields. In Section~\ref{sec:methods}, we introduce the CT-GNN decoder head and how to construct the adjacency matrix. In Section~\ref{sec:eval}, we compare the CT-GNN against other MTC methods on the Sewer-ML dataset, investigate per-class performances, and conduct ablation studies. Finally, in Section~\ref{sec:conclusion}, we conclude the paper.
\section{Related Works}\label{sec:related}

\textbf{Automated Sewer Inspections.} The field of automated sewer inspections has been researched for several decades by both academia and industrial research and development \cite{HaurumSurvey2020}. However, until the release of the Sewer-ML dataset \cite{HaurumDataset2021} there was no public dataset or commonly agreed upon evaluation protocol \cite{HaurumSurvey2020}.

The majority of work within the field has instead focused on automatically classifying defects using CCTV images \cite{HaurumDataset2021} and other sensor based approaches \cite{Alejo2017, Alejo2020,  Bahnsen_2021, SynthPointClouds2, SynthPointClouds, Iyer2012, Khan2018_2,Khan2018_1, Lepot2017, Tezerjani2015}. Only within recent years \cite{HaurumSurvey2020} have deep learning based methods been utilized for defect classification \cite{Chen2018, Hassan2019AlexNet,HaurumDataset2021,  Kumar2018EnsembleCNN,Li2019HierCNN, Meijer2019MultiLabel,  Xie2019HierCNN}, detection \cite{Cheng18,KumarWang20,  Yin20}, segmentation \cite{Kunzel2018, Pan20, Piciarelli19, Wang20}, and spatiotemporal based analysis \cite{Moradi20, WangKumar21}. Defect classification models often employ a two-stage approach with a small initial classifier making a binary defect/non-defect classification, followed by a specialized defect classifier \cite{Chen2018, HaurumDataset2021, Kumar2018EnsembleCNN, Li2019HierCNN,  Xie2019HierCNN}. Recently, work has been conducted on classifying the water level in sewer pipes \cite{HaurumWater2020, Ji2020}, such that it is possible to estimate how much of the pipe can be inspected for defects. However, no work has been conducted on classifying the sewer pipe defects and properties concurrently. For an in-depth review of the vision-based automated sewer inspection field we refer to the survey by Haurum and Moeslund \cite{HaurumSurvey2020}.\\

\textbf{Multi-Task Learning.} The field of multi-task learning has been applied across several different domains. Within the computer vision domain, MTL has been applied on image-level classification tasks such as facial attributes \cite{CelebA2015} and age and gender estimation \cite{imdwiki2015,imdbwiki2018}, learning several unrelated datasets at a time \cite{Omniglot2015, VisDeca2017}, as well as learning multiple dense vision tasks such as per-pixel depth estimation and semantic segmentation \cite{Context2014, Pascal2009, NYUv22012, taskonomy2018}. Two main research branches have been developed through the years: optimization-focused and model-focused approaches \cite{Vandenhende2021MTL}. For an exhaustive review of the field we refer to the surveys of the field \cite{ruder2017MTL, Vandenhende2021MTL, zhang2018MTL}.

The optimization-focused approaches investigate the effect of balancing how the tasks are learned. The tasks are balanced through operations such as normalizing the gradient magnitudes \cite{gradnorm2018}, approaching the problem as a multi-objective optimization problem and finding a Pareto optimal solution among all tasks \cite{Pareto2019, MDGA2018}, adjusting the task weights based on the loss descent rate \cite{MTAN2019}, the task-dependent homoscedastic uncertainties \cite{Uncertainty2018, Uncertainty22018}, and more \cite{GradSign2020, DTP2018, GradSurgery2020}. Each of these approaches is built on different underlying assumptions regarding how the task balancing is controlled, and introduces either an extra computational load or extra hyperparameters.

The model-focused approaches investigate the effect of parameter sharing in the model and is classically split into two types, hard and soft parameter sharing. Hard parameter sharing approaches are built around a shared backbone split into task-specific branches and heads \cite{AutoBranch2020, LearnBranch2020, FAFS2017, GroupTasks2020, Branched2020}, whereas in soft parameter sharing each task is assigned its own parameters with cross-task information introduced through one or more feature sharing mechanisms \cite{NDDRCNN2019, MTAN2019, CrossStitch2016, Sluice2019}. Typically, these models utilize an encoder-decoder structure, where an input is passed through an encoder generating a global or per-task feature representation, which is used by a decoder to produce the task predictions \cite{Vandenhende2021MTL}. This has led to encoder- and decoder-focused methods.

In encoder-focused models the task parameters are only shared in the encoder, while the decoder consists of disjointed task-heads with no cross-task information \cite{gradnorm2018, Uncertainty2018, MDGA2018}. In decoder-focused models, the model parameters are also shared across tasks in the decoder through mechanisms such as multi-model distillation \cite{TRPA2021, MTI2020, PAD2018, PAP2019, PSD2020}, sequential task prediction \cite{JTRL2018}, or cross-task consistency \cite{CrossTask2020}. Decoder-focused models have been applied primarily for dense vision tasks. The few decoder-focused models that have been applied to multi-task classification depend on tensor factorization over pre-trained single task networks \cite{TensorFact2017}, placing a tensor normal prior over the decoder \cite{Multilinear2017} and utilizing a maximum a posteriori optimization objective, or constraining the decoder layers based on the task relations \cite{TaskRelation2019}. However, the previous methods suffer from either requiring initially training single task networks \cite{TensorFact2017}, modifying the optimization loop \cite{Multilinear2017}, or limited to two tasks \cite{TaskRelation2019}. 

Lastly, graphs have seen recent usage in the MTL and MTC fields in modeling between- and within-task relationships. An example of this is the PSD-Net which utilized graphlets to improve per-pixel predictions \cite{PSD2020}. For multi-task classification, graph neural networks (GNNs) have been used to model the relationship between the multiple inputs in a batch \cite{AugFeat2020, RelAtt2018}, or across sequential data \cite{MTLSeq2019}. In concurrent work \cite{LapGraphMTC} a Laplacian graph across facial attributes is learned and used within a regularization term during optimization.

Overall, the literature on MTC decoder-focused models is scarce and existing methods either rely on compressing single task networks or constrained to two tasks. Here, we present a novel decoder-focused model, CT-GNN, which is end-to-end trainable for any number of tasks. Furthermore, in contrast to previous usage of graphs in MTC, the CT-GNN is trainable without relying on sequential or batched data for the graph construction.
\section{Methodology}\label{sec:methods}
In this section, we present our proposed Cross-Task GNN Decoder for Multi-Task Classification. First we provide a recap of Multi-Task Learning and Graph Neural Networks, followed by an explanation of the CT-GNN decoder and how the graph adjacency matrix can be constructed in a data-driven manner.

\subsection{Multi-Task Learning Recap}
Multi-Task Learning focuses on the problem of classifying a set of $T$ tasks, $\mathcal{T}$, simultaneously. Each task contains a set of $C_t$ classes, for a total of $C = \sum_t C_t$ classes. In the case of sewer inspection each image, $\mathcal{I}$, has $T$ task-specific labels $\mathbf{y}_t$. The MTL networks are optimized using a linear combination of the task-specific losses:
\begin{equation}\label{eq:MTL_Loss}
    \mathcal{L}_{\text{Total}} = \sum_{t=1}^T \lambda_t \mathcal{L}_t(\mathcal{I}, \mathbf{y}_t),
\end{equation}
where $\lambda_t$ and $\mathcal{L}_t$ are the weight and loss of the $t$th task, respectively.

When applying multi-task learning methods there are typically varying degrees of parameter sharing in the encoder and no parameter sharing in the decoder. An input image is processed by an encoder network, $f_{\text{ENC}}$, and a set of per-task features $\mathbf{x}_{t} \in \mathbb{R}^{d_{\text{ENC}}}$  are extracted. If there are no task-specific parameters in $f_{\text{ENC}}$ all $T$ tasks will use the same encoded feature $\mathbf{x} \in \mathbb{R}^{d_{\text{ENC}}}$. The encoder features are processed by a decoder network, $f_{\text{DEC}}$, producing predictions for each of the tasks, $\tilde{\mathbf{y}}_t  \in \mathbb{R}^{C_t}$. Classically, $f_{\text{DEC}}$ is constructed as $T$ disjointed classifiers.

\subsection{Graph Neural Network Recap}
A graph, $\mathcal{G} = (\mathcal{V}, \mathcal{E})$, is defined as a set of nodes, $\mathcal{V}$, and edges connecting two nodes, $\mathcal{E}$, together with a set of $d$-dimensional node features $\mathbf{X} \in \mathbb{R}^{|\mathcal{V}|\times d}$. A graph can be represented using an adjacency matrix $\mathbf{A} \in \mathbb{R}^{|\mathcal{V}|\times |\mathcal{V}|}$, where entry $\mathbf{A}[u, v]$ is the edge weight from node $v$ to $u$.  
The basic GNN is defined by its neural message passing structure where the feature vectors of the nodes are exchanged and updated, constituting a GNN layer \cite{GRL2020}. The neural message passing structure for node $u$ and its neighbors $\mathcal{N}(u)$ is defined as:
\begin{equation}\label{eq:gnn}
    \mathbf{h}^{(l+1)}_{u} = \psi(\mathbf{h}^{(l)}_{u}, \phi(\{\mathbf{h}^{(l)}_{v},  \forall v \in \mathcal{N}(u)\})),
\end{equation}
where $\psi$ and $\phi$ are arbitrary differentiable \textit{update} and \textit{aggregation} functions, respectively, and $\mathbf{h}^{l}_{u}$ is the hidden embedding of node $u$ at layer $l$ with $\mathbf{h}^{0}_{u} = \mathbf{x}_{u}$.

\subsection{CT-GNN Decoder for Multi-Task Classification}\label{sec:ctgnn}
The Cross-Task GNN Decoder builds upon the encoder features, $\mathbf{x}_t$, and consists of the following four parts illustrated in Figure~\ref{fig:modelOverview}: $T$ task-specific decoder heads producing the initial per-task feature representations, $T$ bottleneck layers reducing the dimensionality of the per-task feature vectors, $C$ non-linear node embedding layers, and a cross-task GNN which jointly refines the different class representations based on an a priori or learned directed graph $\mathcal{G}_{\mathcal{T}}$.

\textbf{Task-Specific Decoders.} The task-specific decoder heads are realized as a set of $T$ disjointed networks, $f_{\text{DEC}, t}$, each generating a task-specific feature vectors $\mathbf{z}_t = f_{\text{DEC},t}(\mathbf{x}_t)$,  $\mathbf{z}_t\in \mathbb{R}^{d_{\text{DEC}}}$.
Classically, $\mathbf{z}_t$ is used directly to obtain the class predictions, $\check{\mathbf{y}}_t$, by applying a linear layer followed by the classification activation function of choice. In the CT-GNN decoder framework, however, the task-feature $\mathbf{z}_t$ is used as the foundation for the class-specific node embeddings, in order to allow for initial task-adaption of the encoder feature, $\mathbf{x}_t$. 

\textbf{Bottleneck Layer.} In previous work, the dimensionality of the task-specific feature representation $\mathbf{z}_t$ is equal to that of the encoder feature, meaning $d_{\text{ENC}} = d_{\text{DEC}}$ \cite{LearnBranch2020, MDGA2018}. In the CT-GNN decoder framework this is problematic, as the model parameter count would increase dramatically when transforming the $T$ task-specific features into $C$ unique class-specific features of size $d_{\text{EMB}}$. Therefore, a non-linear down projection layer, $f_{\text{BTL},t}$, is applied in order to reduce the dimensionality of the task-specific features and generate a more compact feature representation, $\mathbf{\tilde{z}}_t\in \mathbb{R}^{d_{\text{BTL}}}$. The bottleneck is realized as a dense layer, $\mathbf{\tilde{z}}_t = f_{\text{BTL},t}(\mathbf{z}_t) = \sigma(\mathbf{z}_t \mathbf{B}_t)$, consisting of the down projection weight matrix, $\mathbf{B}_{t}\in \mathbb{R}^{d_{\text{DEC}} \times d_{\text{BTL}}}$, where $d_{\text{BTL}} \leq d_{\text{EMB}} \leq d_{\text{DEC}}$, and applying a differentiable non-linear function, $\sigma$. $\mathbf{B}_t$ can be task-specific or shared across all $T$ tasks, depending on the number of tasks. For a large number of tasks, using task-specific bottleneck layers would result in a large parameter increase, decreasing the parameter-wise benefits of using a MTL network.

\textbf{Node Embeddings.} The dimensionality-reduced task feature representation, $\mathbf{\tilde{z}}_t$, is subsequently turned into $C_t$ class-specific node embeddings. $\mathbf{\bar{z}}_{t,c} \in \mathbb{R}^{d_{\text{EMB}}}$. Similar to the bottleneck layer, this is realized as a dense layer, $\mathbf{\bar{z}}_{t,c} = f_{\text{EMB},t,c}(\mathbf{\tilde{z}}_t) =\sigma(\mathbf{\tilde{z}}_t \mathbf{E}_{t,c})$, consisting of a matrix multiplication and non-linearity. In order to get the $C_t$ unique node embeddings, we use $C_t$ unique embedding layers, parameterized by $C_t$ unique learnable matrices $\mathbf{E}_{t,c}\in \mathbb{R}^{d_{\text{BTL}} \times d_{\text{EMB}}}$.

\textbf{Cross-Task GNN.}
The stacked initial per-class node embeddings, $\mathbf{\bar{Z}} \in \mathbb{R}^{C \times d_{\text{EMB}}}$, of the cross-task graph, $\mathcal{G}_{\mathcal{T}}$, are refined by passing them through a GNN, $\mathbf{\hat{Z}} = f_{\text{GNN}}(\mathbf{\bar{Z}})$, where $\mathbf{\hat{Z}} \in \mathbb{R}^{C \times d_{\text{EMB}}}$ is the stacked GNN-refined node features. The GNN fundamentally builds upon an adjacency matrix of $\mathcal{G}_\mathcal{T}$, $\mathcal{A} \in \mathbb{R}^{C \times C}$, which can be learned, provided a priori, or obtained by a combination thereof. The GNN propagates the node embeddings through $L$ hidden layers with $d_{\text{EMB}}$ channels, adding contextual information to each node embedding based on its incoming neighbors. 

Each node embedding, $\mathbf{\hat{z}}_{t,c} \in \mathbb{R}^{d_{\text{EMB}}}$, is passed through a class-specific linear projection layer, $\hat{z}_{t,c} = f_{\text{CLS},t,c}(\mathbf{\hat{z}}_{t,c})$, to generate a scalar node embedding for each class. The scalar embeddings, $\hat{z}_{t,c}$, are stacked per-task, and the task-specific activation functions are applied to generate the per-task probability vectors, $\mathbf{\hat{y}}_t$. For multi-label and multi-class classification we use the sigmoid and softmax activation.

\subsection{Adjacency Matrix Construction}\label{sec:adj}

A key part of the CT-GNN Decoder is the construction of the graph, realized by the adjacency matrix $\mathcal{A}$. This adjacency matrix can in theory be arbitrarily set. However, in order to utilize the a priori knowledge of the task relationships, we follow a data-driven approach based on the co-occurrence of the classes. We generalize the graph construction method Chen \etal \cite{MLGCN2019} to the multi-task classification scenario. 

$\mathcal{A}$ consists of several sub-matrices, $\mathcal{A}_{i,j}$, each describing the relationship between the tasks $i$ and $j$. Note that in the case that only binary and multi-class classification tasks are considered, $\mathcal{A}$ will be a directed $T$-partite graph with self-loops. Firstly, the conditional probabilities between the classes in task $i$ and $j$, $\mathbf{P}_{i, j} \in \mathbb{R}^{C_i \times C_j}$,  are calculated based on the co-occurrence matrix between the two tasks, $\mathbf{C}_{i, j} \in \mathbb{R}^{C_i \times C_j}$, see Eq.~\ref{eq:condProb}--\ref{eq:condNorm}. The co-occurrence matrices are calculated using the training splits. We follow the convention that $\mathbf{P}_{i, j}[u, v]$ defines the conditional probability of class $u$ given class $v$.

\begin{equation}\label{eq:condProb}
    \mathbf{P}_{i, j}[u, v] = \frac{\mathbf{C}_{i,j}[u,v]}{N_{v}}
\end{equation}
\begin{equation}\label{eq:condNorm}
    N_{v}  = 
    \begin{cases}
        \mathbf{C}_{i, j}[v, v], & i = j\\
        \sum_{u=1}^{C_i} \mathbf{C}_{i, j}[u, v], & i \neq j
    \end{cases}
\end{equation}

$\mathbf{P}_{i, j}$ is subsequently binarized in order to filter out noisy edges using a task-pair specific threshold $\tau_{i,j}$, see Eq.~\ref{eq:binary}. By utilizing task-pair specific thresholds the different task-pairs can be binarized according to different rules, if desired. The binarized adjacency matrices are then combined into a single adjacency matrix, $\mathbf{A}$, see Eq.~\ref{eq:binaryAll}.

\begin{equation}\label{eq:binary}
    \mathbf{A}_{i, j}[u,v]  = 
    \begin{cases}
        0, & \mathbf{P}_{i, j}[u, v] < \tau_{i,j}\\
        1, & \mathbf{P}_{i, j}[u, v] \geq \tau_{i,j}
    \end{cases}
\end{equation}
\begin{equation}\label{eq:binaryAll}
\mathbf{A} = \left[ 
\begin{array}{lll} 
    \mathbf{A}_{1,1} & \dots & \mathbf{A}_{1, K}  \\
    \vdots & \ddots & \vdots  \\
    \mathbf{A}_{K,1} & \dots & \mathbf{A}_{K, K}  \\
\end{array} 
\right] 
\end{equation}

Lastly, the adjacency matrix is re-weighted across the incoming edges per node, in order to counteract the oversmoothing problem with GNNs \cite{MLGCN2019}, leading to the final adjacency matrix, $\mathcal{A}$. This is done using $\mathbf{A}$, and enforcing the sum of all incoming edge weights to equal one, setting the sum of the neighbor edge weights to $p$, while the center node self-loop weight is $1-p$, see Eq.~\ref{eq:reweighted}.

\begin{equation}\label{eq:reweighted}
    \mathcal{A}[u,v]  = 
    \begin{cases}
        \mathbf{A}[u,v]\frac{p}{\sum_{v=1, v\neq u }^{C_j}\mathbf{A}[u,v]}, & u \neq v\\
        1-p, & u = v
    \end{cases}
\end{equation}

The larger $p$ is the more weight will be assigned to the incoming neighbor nodes, while a smaller $p$ value will result in more weight assigned to the center node. If a center node has no incoming edges a part from the self-loop, \ie $\sum_{v=1, v\neq u }^{C_j}\mathbf{A}[u,v] = 0$, we set the self-loop weight to one, to avoid the center node embedding decaying to a zero vector.

\section{Experimental Results}\label{sec:eval}
We evaluate on the Sewer-ML sewer defect and pipe property dataset \cite{HaurumDataset2021}. 
The dataset focuses on the multi-label defect classification problem and contains 1.3 million images collected over a nine year period. The data are split into a preset training, validation, and test split, containing 1 million, 130k, and 130k images each \cite{HaurumDataset2021}. The defect classification problem consists of 17 different classes as well as the implicit normal class. Additionally, the water level, pipe material and pipe shape are also annotated. The water level is annotated in 11 classes from 0 to 100\% of the pipe filled with water in 10\% steps, and the pipe material and shape tasks contain eight and six classes each. Example images can be found in the supplementary material.

\subsection{Evaluation Metrics}
Model evaluation is done using the per-task evaluation metrics and number of parameters, \#P. As the classes in each task are imbalanced the tasks cannot be evaluated using the traditional accuracy metric. Instead, the defect task is evaluated using the $\text{F}2_{\text{CIW}}$ defect score and the $\text{F}1_{\text{Normal}}$ score \cite{HaurumDataset2021}. The three remaining tasks are evaluated using both the micro-F1 (mF1) and macro-F1 (MF1) scores. 

 Lastly, we report the average per-task performance increase for a multi-task model, $\Delta_{\text{MTL}}$, with respect to the single task learning (STL) baselines of the same base architecture \cite{AST2019}:
\begin{equation}\label{eq:delta}
    \Delta_{\text{MTL}} = \frac{1}{T} \sum_{t=1}^T \frac{(M_{m,t}-M_{b,t})}{M_{b,t}},
\end{equation}
where $M_{m,t}$ and $M_{b,t}$ are the multi-task and single-task metric performance for task $t$, receptively.

\begin{table}[!t]
\centering
\caption{\textbf{CT-GNN hyperparameters}. The hyperparameters were found through a sequential search. $L$ is the number of layers in the CT-GNN, $d_{\text{ENB}}$ is the dimensionality of the class features, $d_{\text{BTL}}$ is the dimensionality of the bottleneck, $H$ is the number of attention heads in the GAT GNN, and $\tau$ and $p$ are the thresholding and re-weighting parameters in the adjacency matrix construction, respectively.}
\label{tab:sewerHP}
\begin{tabular}{lccccccc}\toprule
Hyperparameter & $L$ & $d_{\text{EMB}}$ & $d_{\text{BTL}}$ &$H$ & $\tau$ & $p$ \\ \cmidrule(r){1-7}
GCN & 3 & 512 & 32 &  -&0.05 & 0.2 \\
GAT & 1 &128 & 32 & 8 & 0.65 & - \\
 \bottomrule
\end{tabular}
\end{table}

\begin{table*}[!t]
\centering
\caption{\textbf{Results on Sewer-ML.} Comparison between the STL and MTL networks. We compare the effect of CT-GNN using GCN and GAT, denoted CT-GCN and CT-GAT respectively, as well as compare a hard-shared ResNet-50 encoder, and the soft-shared MTAN encoder with a ResNet-50 backbone. \#P indicates the number of parameters in millions. * indicates that the method was tested on a subset of the Sewer-ML dataset. Best performance in each column is denoted in \textbf{bold}.}
\label{tab:sewerRes}
\begin{tabular}{llccrrccccccc} \toprule
&\multicolumn{2}{c}{Model} & \multicolumn{1}{c}{Overall} & \multicolumn{2}{c}{Defect} & \multicolumn{2}{c}{Water} & \multicolumn{2}{c}{Shape} & \multicolumn{2}{c}{Material}  \\ 
\cmidrule(r){2-3} \cmidrule(r){4-4} \cmidrule(r){5-6} \cmidrule(r){7-8} \cmidrule(r){9-10} \cmidrule(r){11-12} 

& Model & \#P  & \textbf{$\Delta_{\text{MTL}}$}& \textbf{$\text{F}2_{\text{CIW}}$} & \textbf{$\text{F}1_{\text{Normal}}$} & \textbf{MF1} & \textbf{mF1} & \textbf{MF1} & \textbf{mF1} & \textbf{MF1} & \textbf{mF1} \\ \cmidrule(r){1-12} 

\parbox[t]{2mm}{\multirow{7}{*}{\rotatebox[origin=c]{90}{Validation Split}}} &  Benchmark \cite{HaurumDataset2021}  & 62.8& -& 55.36 & 91.32 & - & - & - & - & - & -    \\ 
& R50-FT* \cite{HaurumWater2020}  & 23.5& -& - & - & 62.53 & 78.15 & - & - & - & -    \\ 

\cmidrule(r){2-12} 

& STL & 94.0& +0.00 & 58.42 & \textbf{92.42} & 69.11 & 79.71 & 46.55 & 98.06 & 65.99 & 96.71   \\

& R50-MTL & 23.5  & +10.36 & 59.73&	91.87&	70.51&	80.47 &	71.64&	99.34&	80.28&	98.09\\

& MTAN & 48.2 & +10.40 &61.21&	92.10&	70.06&	\textbf{80.59}&	68.34&	\textbf{99.40}&	83.48&	\textbf{98.25}  \\

\cmidrule(r){2-12} 
& CT-GCN & 25.2 & +12.39 & 61.35 & 91.84 & \textbf{70.57} &80.47 & \textbf{76.17} & 99.33	& 82.63	&98.18   \\ 
& CT-GAT & 24.0& +\textbf{12.81} & \textbf{61.70} &	91.94&	\textbf{70.57} &	80.43&	74.53&	\textbf{99.40}&	\textbf{86.63}&	98.24    \\

\cmidrule(r){1-12}\morecmidrules\cmidrule(r){1-12}

\parbox[t]{2mm}{\multirow{7}{*}{\rotatebox[origin=c]{90}{Test Split}}} &  Benchmark \cite{HaurumDataset2021} & 62.8& - & 55.11 & 90.94  & - & - & - & - & - & -    \\ 
& R50-FT* \cite{HaurumWater2020} & 23.5& -& - & - & 62.88 & 79.29 & - & - & - & -    \\ 

\cmidrule(r){2-12} 

&STL & 94.0 & +0.00 & 57.48 & \textbf{92.16} & 69.87 & 80.09 & 56.15 & 97.59 & 69.02 & 96.67   \\

& R50-MTL & 23.5& +7.39  & 58.29&	91.57&	71.17&	81.09&	79.48&	99.19&	\textbf{76.35}&	98.08   \\
& MTAN& 48.2& +6.83 & 59.91&	91.72&	70.61&	\textbf{81.16}&	78.50&	99.21&	72.73&	\textbf{98.27}   \\

\cmidrule(r){2-12} 
&CT-GCN& 25.2& +7.64 &  60.07&	91.60&	70.69&	80.91&	80.32&	99.19&	75.13&	98.15    \\ 
&CT-GAT & 24.0 & +\textbf{7.84} & \textbf{60.57}&	91.61&	\textbf{71.30}&	80.91&	\textbf{81.10}&	\textbf{99.22}&	73.95&	98.26   \\

\bottomrule

\end{tabular}
\end{table*}
\subsection{Training Procedure}
\newlength{\oldintextsep}
\setlength{\oldintextsep}{\intextsep}

We utilize the ResNet-50 network \cite{ResNet} as our base encoder, with no task-specific decoders, meaning $\mathbf{x}_t = \mathbf{z}_t$. We cast the defect classification problem as a multi-label classification task with a single task weight, $\lambda_{\text{defect}}$, while the water level, pipe material, and pipe shape are multi-class classification tasks. For the water level classification task, we adapt the label discretization approach from \cite{HaurumWater2020}, leading to four water level classes.

We compare performance using the Graph Convolutional Network (GCN) \cite{GCN2017} and Graph Attention Network (GAT) \cite{GAT2018} in the CT-GNN, denoting the variations CT-GCN and CT-GAT, respectively.  
We use the reweighted adjacency matrix, $\mathcal{A}$, for GCN, and the binary adjacency matrix, $\mathbf{A}$, for GAT where the edge weights are inferred through self-attention. While the GAT architecture could fully determine the adjacency matrix through self-attention, we found that performance increases if we provide the set of possible graph edges beforehand. The GCN adjacency matrix was symmetrically normalized \cite{GCN2017} using the in-degree matrix, and skip connections were inserted between the GNN layers. Finally, we use task-specific bottleneck layers.

\textbf{Hyperparameters.} The networks are trained for 40 epochs using SGD with a learning rate of 0.1, momentum of 0.9, weight decay of 0.0001, and a batch size of 256. The learning rate is multiplied by 0.01 at the 20th and 30th epoch. The hyperparameters used in the CT-GNN, including the number of attention heads in GAT, $H$, are described in Table~\ref{tab:sewerHP}, and are found through a sequential hyperparameter search described in the supplementary material. Through initial tests we found that a single global threshold $\tau$ in the adjacency graph construction leads to the best performance. 

\textbf{Data Augmentation.} We follow the data augmentation process by \cite{HaurumDataset2021}, rescaling the images to $224\times 224$, horizontal flipping and jittering the brightness, contrast, hue, and saturation values by $\pm10\%$. Due to class imbalance in each task, we use class-weighted task-losses with the class weighting method of \cite{Cui2019} with $\beta=0.9999$, except for the defect task where the positive class examples are weighted by their \textit{class importance weights} (CIW) \cite{HaurumDataset2021}.

\textbf{Loss considerations.} For all CT-GNN models the final task loss is a convex combination of the final probability vector $\mathbf{\hat{y}}_t$ and the probability vector produced by applying a classification layer to $\mathbf{z_t}$, denoted $\mathbf{\check{y}}_t$:
\begin{equation}\label{eq:auxLoss}
    \mathcal{L}_{t} = \omega \mathcal{L}_{t}(\hat{\mathbf{y}}_t, \mathbf{y}_t) + (1-\omega)\mathcal{L}_{t}(\check{\mathbf{y}}_t, \mathbf{y}_t),
\end{equation}
where $\mathcal{L}_t$ is the task-specific loss function for task $t$, and $\omega$ is a weighting hyperparameter in the interval $[0,1]$.
This is to ensure the feature representation $\mathbf{z_t}$ is representative for task $t$, through an auxiliary loss signal.  We set $\omega = 0.75$, such that the primary loss signal is propagated through the CT-GNN.\\

We constrain the task weights to be a convex combination and set to $\lambda_{\text{defect}} = 0.90$ and $\lambda_{\text{water}} = \lambda_{\text{shape}} =\lambda_{\text{material}} = \frac{1-\lambda_{\text{defect}}}{3}$.
In order to keep the losses comparable across different settings, we multiply the task weights by $T$ such that $\sum_t \lambda_t = T$, similar to \cite{MTAN2019}.

\subsection{Comparative Models}

As there are no ResNet-50 STL baselines for all of the tasks, we train these using the same hyperparameters as the in MTL networks. Note that we got the best single-task performance for the defect task using the class weighting method from \cite{Cui2019}. We also compare with the benchmark defect classification model from \cite{HaurumDataset2021}, as well as the water level classification model from \cite{HaurumWater2020}. As there are no prior work on multi-task classification in the sewer domain \cite{HaurumSurvey2020}, we compare with a set of MTL baselines: A hard-shared ResNet-50 MTL network with no CT-GNN (R50-MTL), and the encoder-focused soft-shared MTAN model with a ResNet-50 backbone, see Table~\ref{tab:sewerRes}. Results for the DWA \cite{MTAN2019} and the uncertainty \cite{Uncertainty2018,Uncertainty22018} optimization-based methods can be found in the supplementary materials. 

\begin{figure}[!t]
    \centering
    \includegraphics[width=0.78\linewidth]{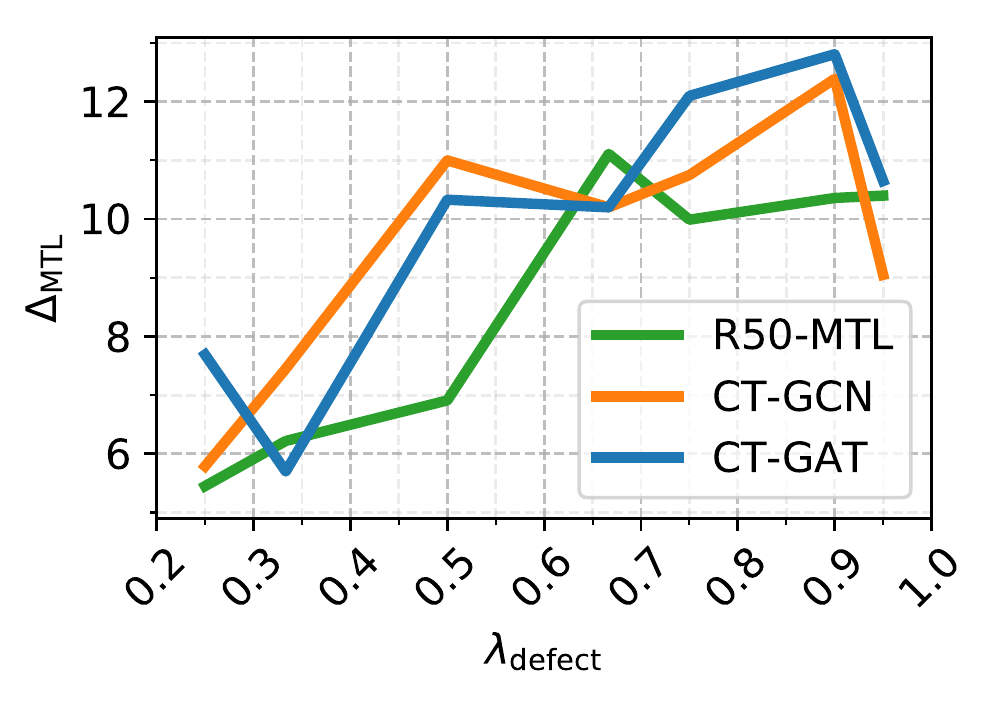}
    \caption{\textbf{Evaluating $\Delta_\text{MTL}$ for different $\lambda_\text{defect}$.}  Comparison of performance of the R50-MTL, CT-GCN and CT-GAT models. Evaluated on the validation split.}
    \label{fig:abl_taskWeight}
\end{figure}

\subsection{Results}

We find that the CT-GNN outperforms all other methods, beating state-of-the-art defect \cite{HaurumDataset2021} and water level \cite{HaurumWater2020} classifiers by 5.3 and 8.0 percentage points, respectively. We also outperform the baseline STL and MTL networks, by a significant margin on the defect, shape, and material tasks.

The CT-GCN and CT-GAT achieve comparable or better metric performance on all tasks while adding 0.5-1.7 million parameters compared to MTAN encoder-focused method which adds 25 million parameters. Specifically, CT-GAT achieves the highest $\Delta_\text{MTL}$ while introducing 50 times fewer parameters than the MTAN encoder. Unlike soft-shared encoders, the backbone only influences the parameter count of the CT-GNN through the size of the encoder feature $\mathbf{x}_t$.

Comparatively, the optimization-based methods performed worse than using a fixed set of task weights, echoing the results from \cite{Vandenhende2021MTL}, resulting in a $\Delta_\text{MTL}$ of -15.70\% and -4.07\% on the validation split and -11.57\% and -4.07\% on the test split, for the DWA and uncertainty methods respectively. Details are available in the supplementary material.

We also find that the CT-GAT outperforms the CT-GCN on the defect and materials task, while the CT-GCN performs slightly better on the shape task MF1 score. This indicates that there is a clear value in letting the edge weights be dynamically inferred during inference, while prior information can be imbued beforehand through the structure of the adjacency matrix. Furthermore, it demonstrates that good performance can be achieved with limited prior knowledge of the task and class relationships.

Lastly, we observe that the general performance, as measured by $\Delta_{\text{MTL}}$, increases when using MTL networks. By inspecting the results, one can see that the water task performance is not affected by the MTL networks. However, for the defect, material and shape tasks the performance increases dramatically, beating the STL method and benchmark method from \cite{HaurumDataset2021} by several percentage points, indicating a clear benefit of utilizing an MTL approach. We also observe a clear difference in $\Delta_\text{MTL}$ across the validation and test splits. This is attributed to the shape and material tasks where the classes are very imbalanced, leading to few labels to learn from during training and a potentially large difference between the examples in the different splits.

\begin{figure*}[!htp]
     \centering
     \begin{subfigure}[b]{0.34\textwidth}
         \includegraphics[width=\textwidth]{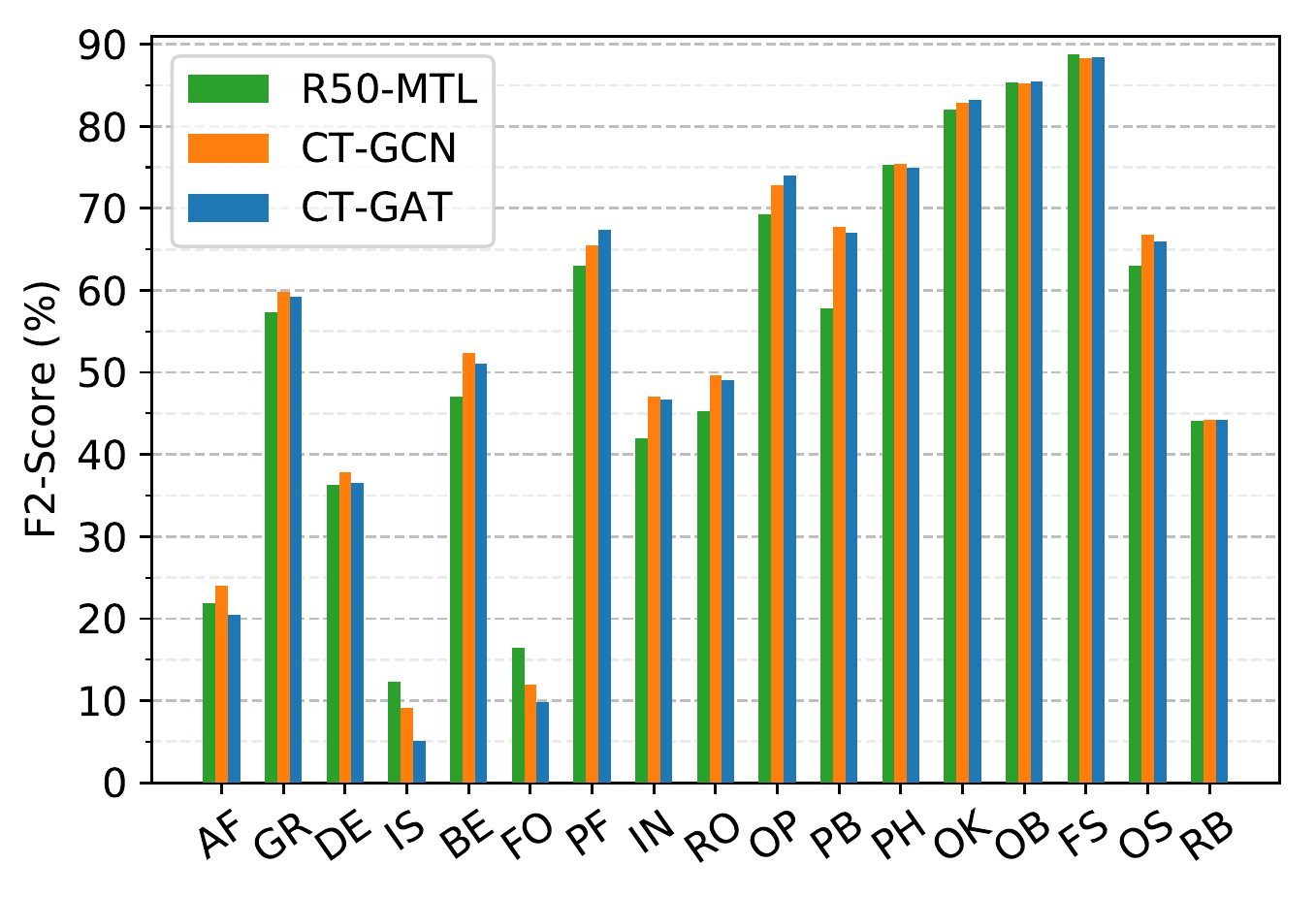}
         \caption{Per-class defect performance}
         \label{fig:defectf2}
     \end{subfigure}
     \begin{subfigure}[b]{0.3\textwidth}
         \includegraphics[width=\textwidth]{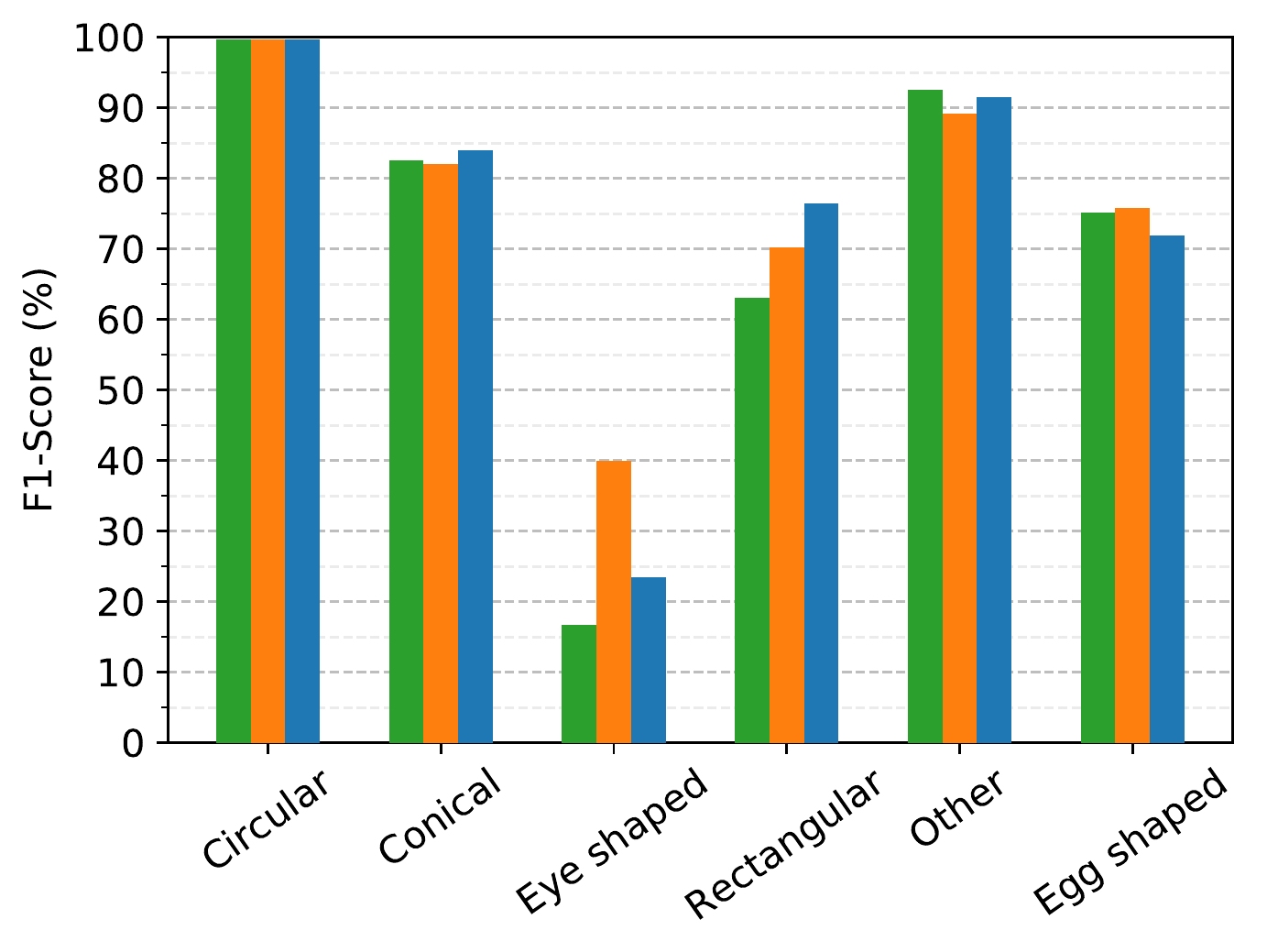}
         \caption{Per-class shape performance}
         \label{fig:shapef1}
     \end{subfigure}
     \hfill
     \begin{subfigure}[b]{0.3\textwidth}
         \includegraphics[width=\textwidth]{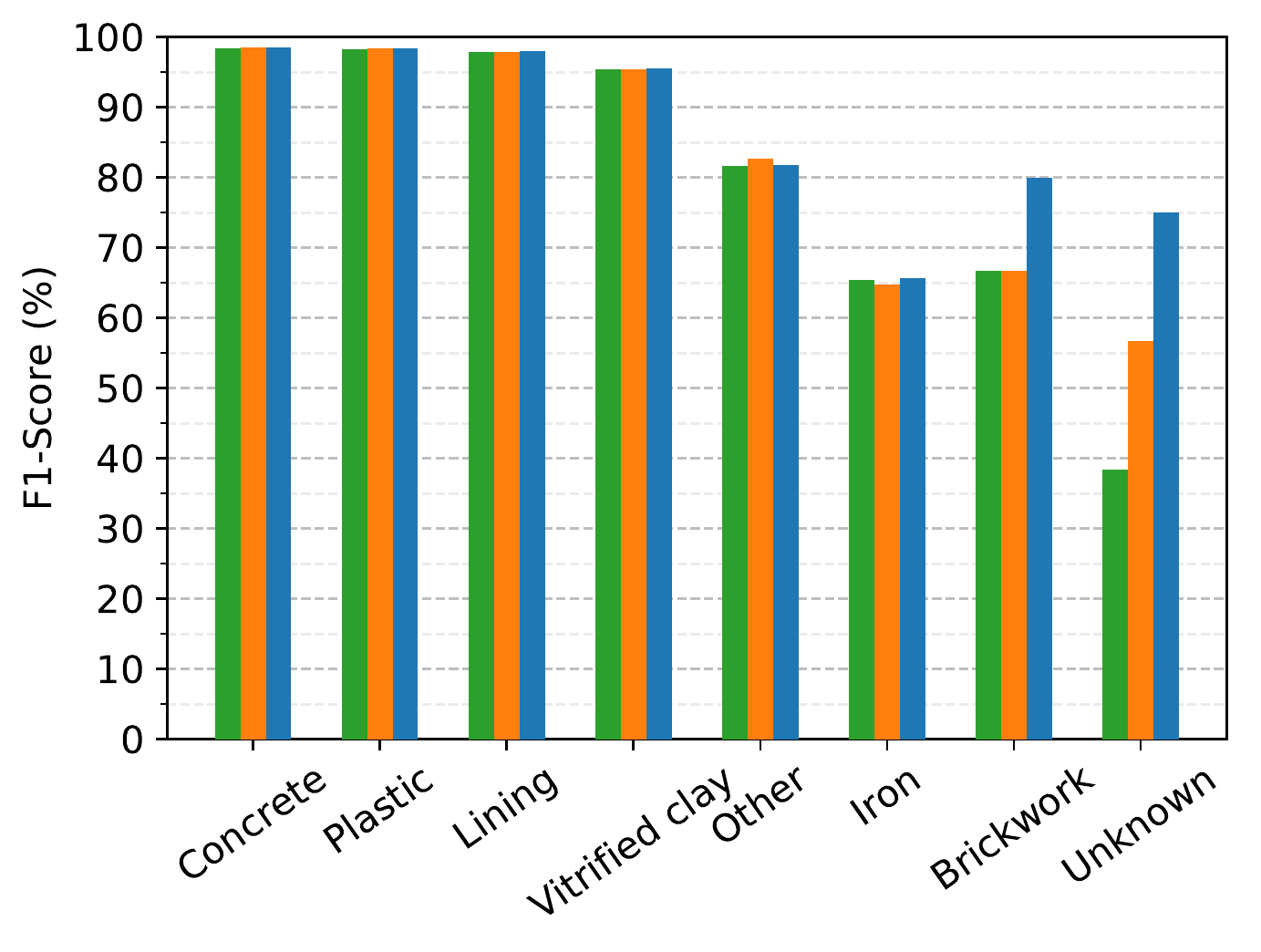}
         \caption{Per-class material performance}
         \label{fig:matf1}
     \end{subfigure}
        \caption{\textbf{Per task class comparisons.} We compare model performance on the validation set. The F2 defect scores are plotted for each defect class in Figure~\ref{fig:defectf2} ordered by increasing CIW from left to right. We refer to the Sewer-ML paper \cite{HaurumDataset2021} for an explanation of the defect class codes. The class F1-scores for the shape and material tasks are plotted in Figure~\ref{fig:shapef1}-\ref{fig:matf1}. The scores are plotted by decreasing number of training samples per class.}
        \label{fig:three graphs}
\end{figure*}
\subsection{Per-task Analysis of Results}
To get a better understanding of the performance difference between the CT-GNNs and R50-MTL, we dive into the per-class task performances. Images of the different classes can be found in the supplementary material.

When comparing the individual defect F2-scores, shown in Figure~\ref{fig:defectf2}, we see that the CT-GNN performs better on defects with high CIWs but few training examples such as OS, PB, and PS, while the performance is worse on the rare defect classes with a low CIW such as IS and FO. For classes where there are plenty of examples to learn from we observe that the performance is comparable across all models.

When investigating the water task, we observe that all models perform equally well on all classes. On the shape task it is clear the CT-GNN performs better on the rectangular and eye shaped pipes, see Figure~\ref{fig:shapef1}. It should be noted that the amount of validation examples of eye shaped pipes is very low. The CT-GNN does, however, achieve a slightly lower F1-score on the egg shaped pipes. On the material task, the CT-GNN again improves performance compared to the baseline, see Figure~\ref{fig:matf1}. By using the CT-GAT performance on the Brickwork and Unknown classes increase by 13 and 37 percentage points, respectively.

\subsection{Ablation Studies}

\textbf{Importance of $\lambda_\text{defect}$.} The most critical part of an automated sewer inspection system, is the capability to classify the presence of defects correctly.  Therefore, we investigate the effect of different $\lambda_\text{defect}$ values on the overall performance metric $\Delta_\text{MTL}$. We compare the performance when setting $\lambda_\text{defect} = \{0.25, 0.33, 0.50, 0.67, 0.75, 0.90, 0.95\}$ ranging from an equal weighting between all four tasks ($\lambda_\text{defect} = 0.25$) to focusing on the defect task ($\lambda_\text{defect} = 0.95$). We train an MTL model with a hard-shared ResNet-50 encoder with and without the CT-GNN decoder heads, see Figure~\ref{fig:abl_taskWeight}. We observe that the $\Delta_\text{MTL}$ increases steadily together with $\lambda_\text{defect}$, peaking at $\lambda_\text{defect}=0.90$, before decreasing when prioritizing the defect task too much when $\lambda_\text{defect}=0.95$.

\textbf{Combining MTAN and CT-GNN.} The combination of soft parameter sharing encoder- and decoder-focused models has not previously been investigated. Therefore, we compare the effect of combining MTAN encoder and the CT-GNN decoder, to determine whether the two approaches are complementary.
We find that the CT-GCN and CT-GAT obtains a $\Delta_\text{MTL}$ of 12.72\% and 11.48\% when trained with MTAN, respectively. This shows that the combination of MTAN and CT-GCN leads to a higher performance with the CT-GCN compared to using a hard-shared encoder. However, when using the CT-GAT the performance decreases. This indicates the GNN settings cannot just be transferred from a hard to soft-shared encoder, instead requiring a small search over how the graph is constructed. 

\noindent The per-task metric performances for both ablation studies can be found in the supplementary material. 
\section{Conclusion}\label{sec:conclusion}
One of the most important infrastructures in modern society is the sewerage infrastructure, but it is difficult to inspect and maintain. Automated sewer inspection methods have been investigated for decades, with an emphasis of the important defect classification task, while sewer properties such as water level, pipe material, and pipe shape, which are needed to determine the deterioration level, have been neglected.

We approach the automated sewer inspection problem as a multi-task classification problem. To this end we introduce our novel Cross-Task Graph Neural Network (CT-GNN) Decoder, which utilizes the cross-task information between concurrent and related tasks to refine the per-task predictions. This is realized by generating unique per-class node embeddings that are combined and refined through the use of a graph neural network. 

Using our novel method, we not only beat the state-of-the-art on the defect and water level classification tasks by 5.3 and 8.0 percentage points, respectively, but also outperform other single-task and multi-task learning methods on all four classification tasks in the Sewer-ML dataset \cite{HaurumDataset2021}. Furthermore, the CT-GNN decoder introduces 50 times fewer parameters compared to encoder-focused models.

The novel CT-GNN approach is focused on handling the concurrent image-level classification tasks present in the Sewer-ML dataset. It is, however, important to note that the method is not specific to the sewer data and can therefore be expected to generalize to other domains containing concurrent classification tasks. Another interesting future direction for the CT-GNN is to adapt it to regression tasks where the values cannot be discretized.

\ifwacvfinal
\subsection*{Acknowledgments}
This research was funded by Innovation Fund Denmark [grant number 8055-00015A] and is part of the Automated Sewer Inspection Robot (ASIR) project, and partially supported by the Spanish project PID2019-105093GB-I00 (MINECO/FEDER, UE), and by ICREA under the ICREA Academia programme.
\fi

\appendix
\section{Supplementary Materials Content}

In these supplementary materials we describe the hyperparameter search, more in-depth results for the optimization-based multi-task learning (MTL) methods as well as the ablation studies. We also show examples of the different task classes, and show examples of success and failure cases for the CT-GNN. Specifically, the following will be described:
\begin{itemize}
    \itemsep0em
    \item Example images of the different task classes (Section~\ref{sec:taskExamples}).
    \item Hyperparameter search (Section~\ref{sec:HPSearch}).
    \item In-depth optimization-based MTL results (Section~\ref{sec:opt}).
    \item In-depth results for the $\lambda_\text{defect}$ ablation study (Section~\ref{sec:lambda}).
    \item In-depth results for the MTAN and CT-GNN ablation study (Section~\ref{sec:mtan}).
    \item Examples of how the CT-GNN succeeds and fails (Section~\ref{sec:success}).
\end{itemize}

\section{Sewer-ML Task Class Examples}\label{sec:taskExamples}
For the sake of clarity we show examples of each class in the water level, pipe shape and pipe material tasks, see Figure~\ref{tab:waterExamples}-\ref{tab:matExamples}. For examples of the pipe defect classes we refer to the supplementary materials of the Sewer-ML paper \cite{HaurumDataset2021}.

\begin{table}[!h]
\centering
\caption{\textbf{Initial Hyperparameter Values.} The investigated hyperparameters are set to the following starting values, and after each step of the sequential search the corresponding hyperparameter is updated. It should be noted that $\tau$ used in the GAT GNN was set to 0.05. This was done to reduce the amount of noisy graph edges in the Sewer-ML dataset, caused by the large class imbalance in some tasks.}
\label{tab:HPInit}
\begin{tabular}{lcc}\toprule
Hyperparameter & GCN & GAT \\ \cmidrule(r){1-3}
$L$ & 2 & 2 \\
$d_{\text{EMB}}$ & 256 & 256\\
$d_{\text{BTL}}$ & 32 & 32\\
$H$ & - & 8 \\
$\tau$ & 0.05  & 0.05\\
$p$ & 0.2 & - \\
 \bottomrule
\end{tabular}
\end{table}
\section{Hyperparameter Search}\label{sec:HPSearch}
In the hyperparameter search for the CT-GNN decoder we investigated the effect when varying the design of the bottleneck layer and the CT-GNN. The investigated parameters and their search space is presented in Table~\ref{tab:HPInterval}. It should be noted that the amount of attention heads, $H$, and the re-weighting parameter, $p$, were only utilized for the GAT \cite{GAT2018} and GCN \cite{GCN2017} GNNs, respectively.
Due to the amount of hyperparameters and the size of the value ranges, we decided to employ a sequential hyperparameter search design. The search was initialized with the hyperparameters stated in Table~\ref{tab:HPInit}. All tests were performed with $\lambda_\text{defect} = 0.50$ to ensure a fair weighting of the tasks, while prioritizing the defect task.

\begin{table*}[!h]
\centering
\caption{\textbf{Investigated Hyperparameters}. The hyperparameters of the CT-GNN and the Bottleneck layer were investigated. For each hyperparameter we have denoted the values investigated.}
\label{tab:HPInterval}
\begin{tabular}{lc}\toprule
Hyperparameter & Range   \\ \cmidrule(r){1-2}
$L$ & [1, 2, 3]  \\
$d_{\text{EMB}}$& [128, 256, 512] \\
$d_{\text{BTL}}$ & [16, 32, 64, 128]  \\
$H$ & [1, 2, 4, 8, 16]   \\
$\tau$ & [0.00, 0.05, 0.15, 0.25, 0.35, 0.45, 0.55, 0.65, 0.75, 0.85, 0.95] \\
$p$ & [0.1, 0.2, 0.3, 0.4, 0.5, 0.6, 0.7, 0.8, 0.9]   \\
 \bottomrule
\end{tabular}
\end{table*}

At each step of the search the best performing hyperparameter was kept and used for all future steps of the search. The order of the sequential search was realized as follows:

\begin{enumerate}
    \itemsep0em
    \item Grid search across the number of GNN layers, $L$, and the number of GNN channels, $d_{\mathrm{EMB}}$.
    \item Search over the number of channels in the bottleneck layer, $d_{\text{BTL}}$.
    \item Search over the number of attention heads, $H$. \textbf{Only performed for GAT.}
    \item Search over the adjacency matrix threshold, $\tau$.
    \item Search over the adjacency matrix neighbor node reweighting parameter, $p$. \textbf{Only performed for GCN.} 
\end{enumerate}

\begin{figure}[!t]
\centering
  \includegraphics[width=0.65\linewidth]{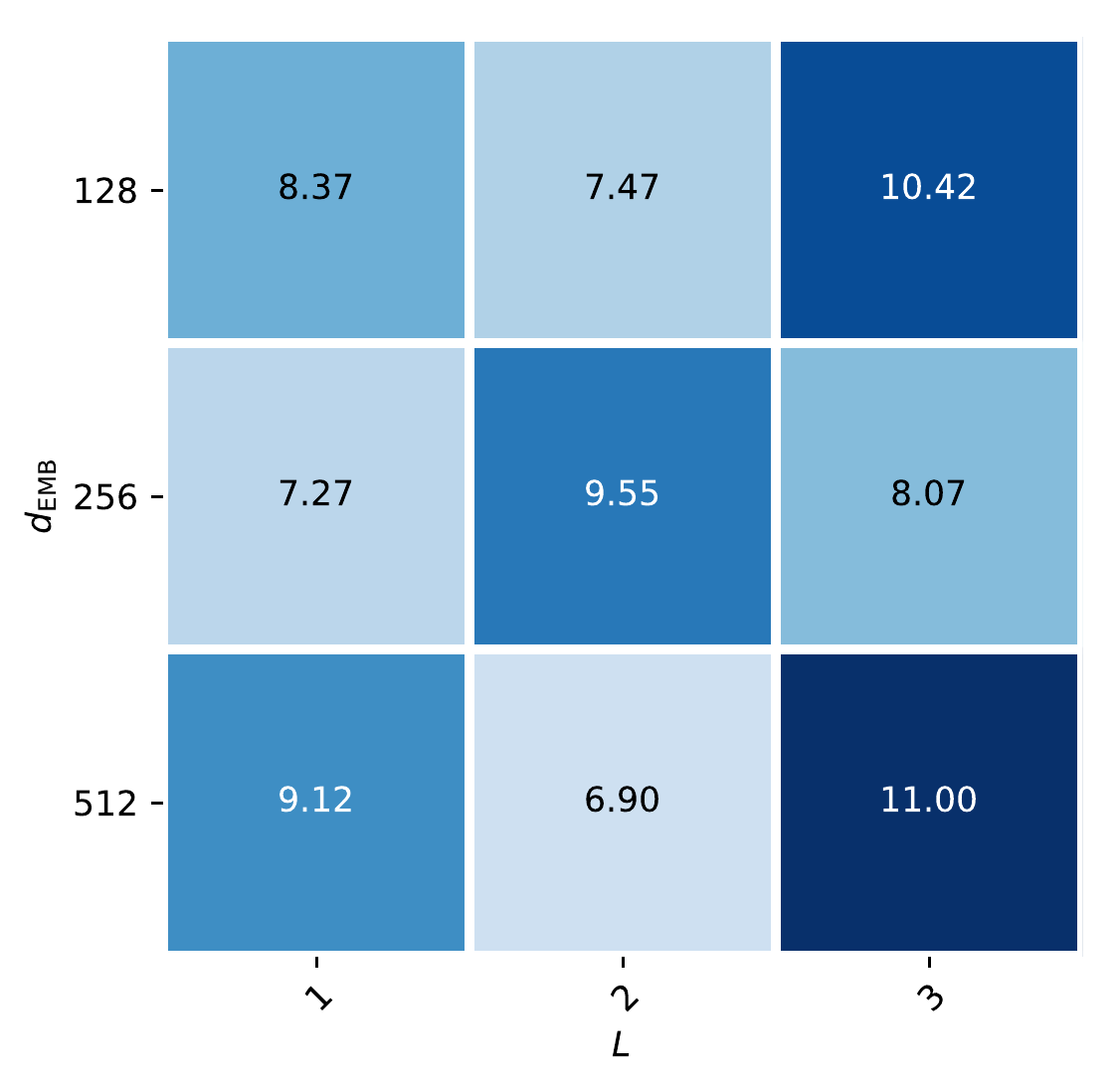}
  \captionof{figure}{Grid search over $L$ and $d_{\text{EMB}}$ for CT-GCN.}
  \label{fig:SewerGCNHeat}
\end{figure}

\begin{figure}[!t]
\centering
  \includegraphics[width=0.65\linewidth]{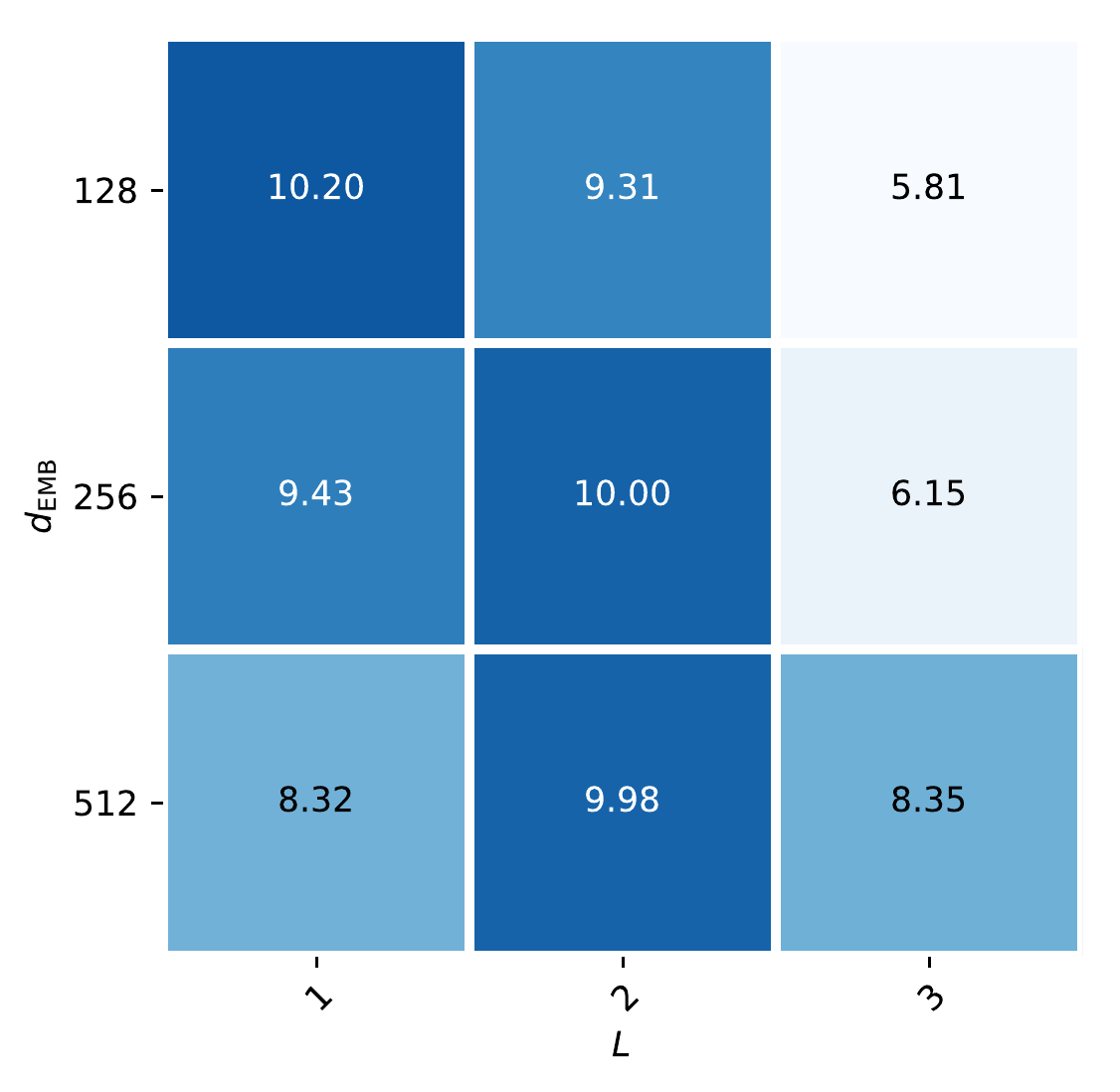}
  \captionof{figure}{Grid search over $L$ and $d_{\text{EMB}}$ for CT-GAT.}
  \label{fig:SewerGATHeat}
\end{figure}

\begin{figure}[!htp]
  \centering
  \includegraphics[width=0.95\linewidth]{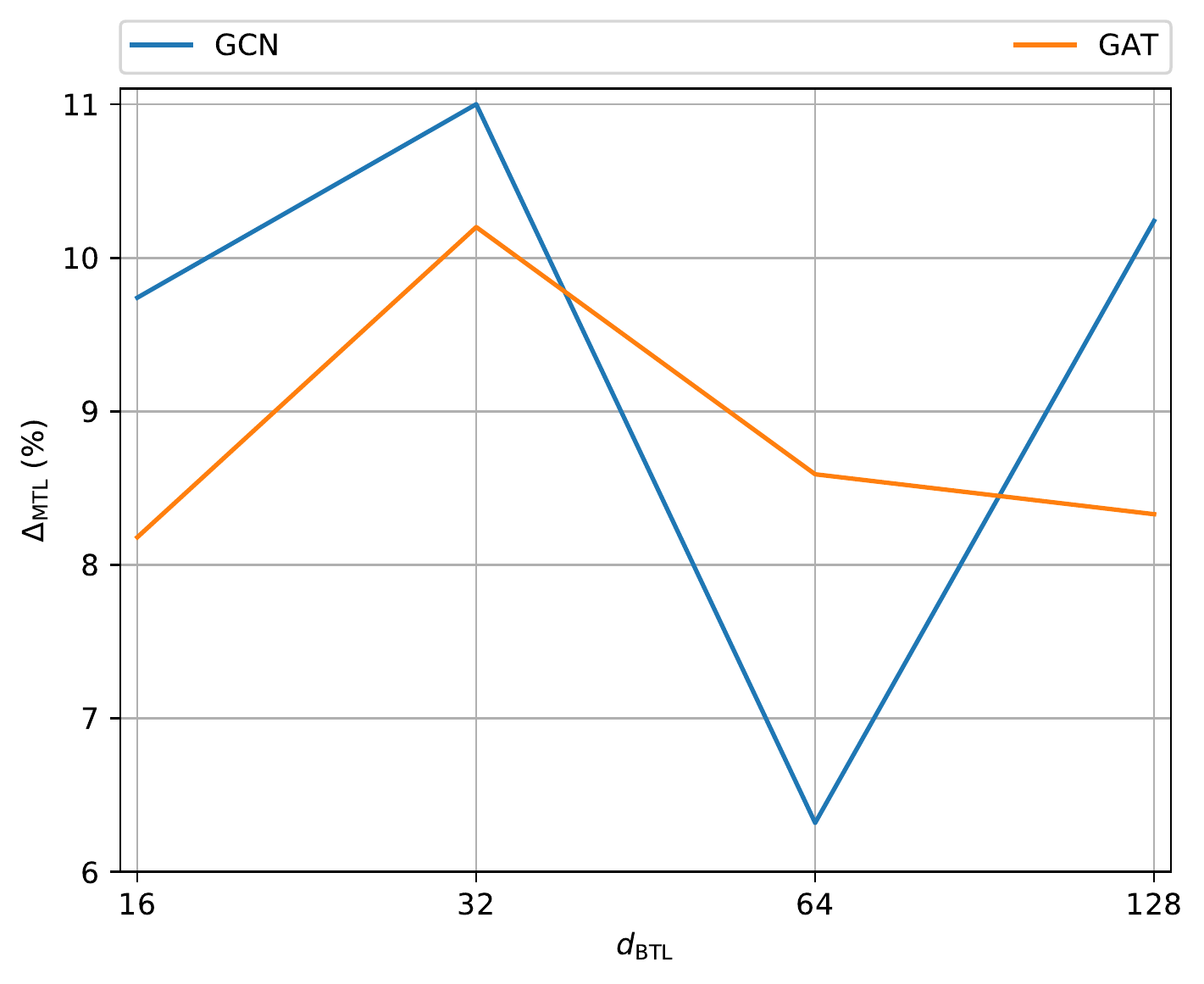}
  \captionof{figure}{Search over $d_{\text{BTL}}$.}
  \label{fig:SewerB}
\end{figure}

\begin{figure}
  \centering
  \includegraphics[width=0.95\linewidth]{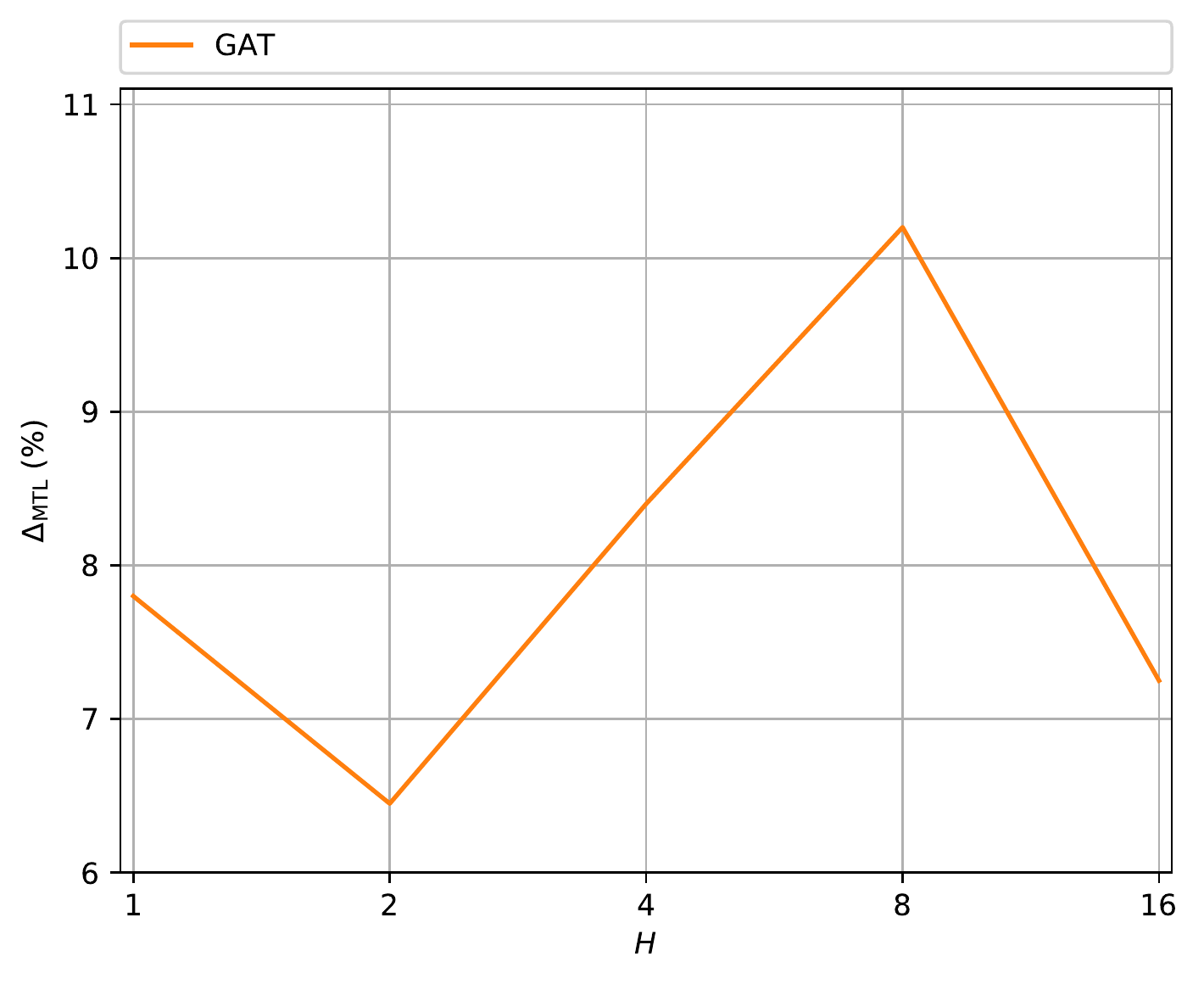}
  \captionof{figure}{Search over $H$.}
  \label{fig:SewerH}
\end{figure}

\begin{figure}[!htp]
  \centering
  \includegraphics[width=0.95\linewidth]{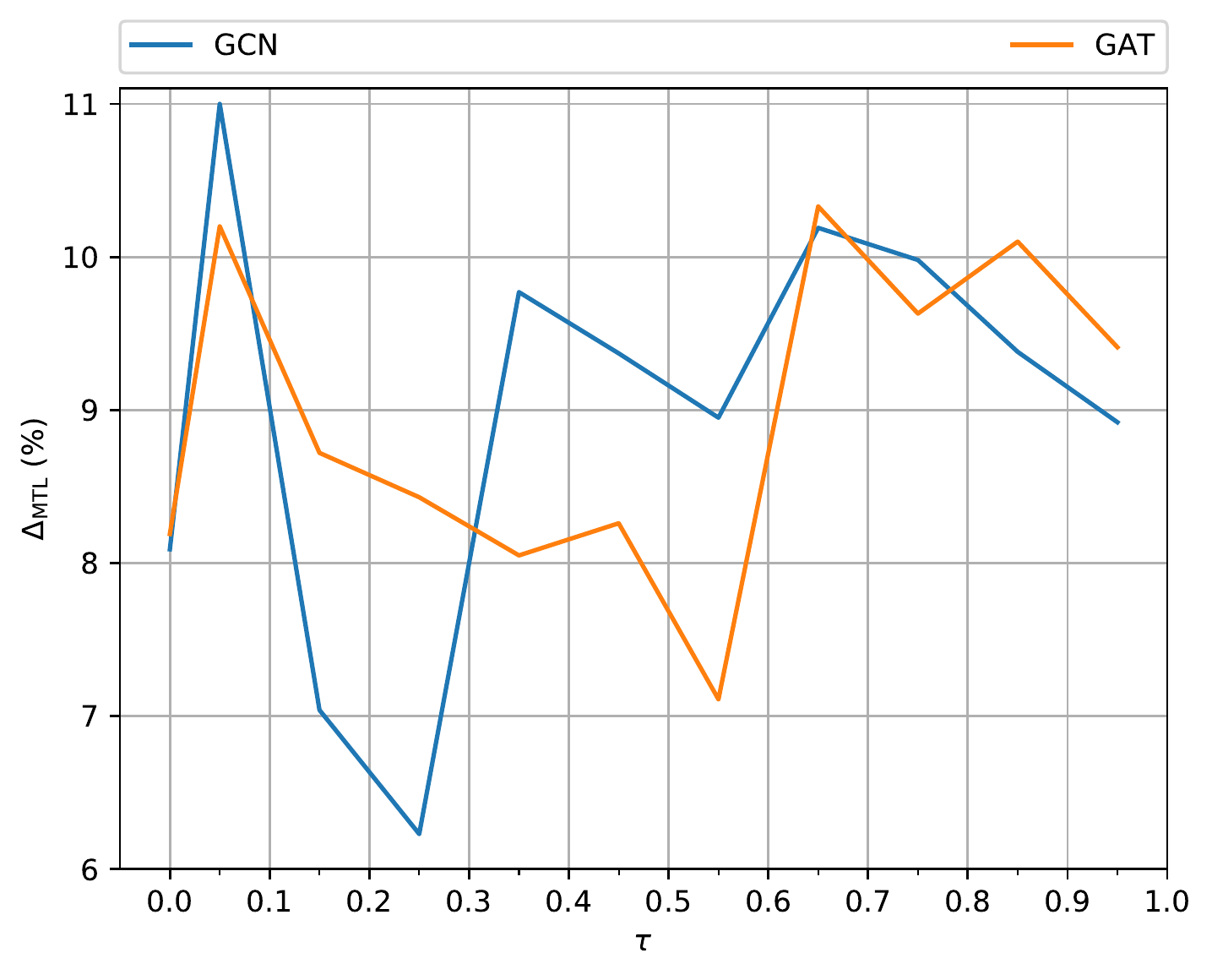}
  \captionof{figure}{Search over $\tau$.}
  \label{fig:SewerT}
\end{figure}

\begin{figure}
  \centering
  \includegraphics[width=0.95\linewidth]{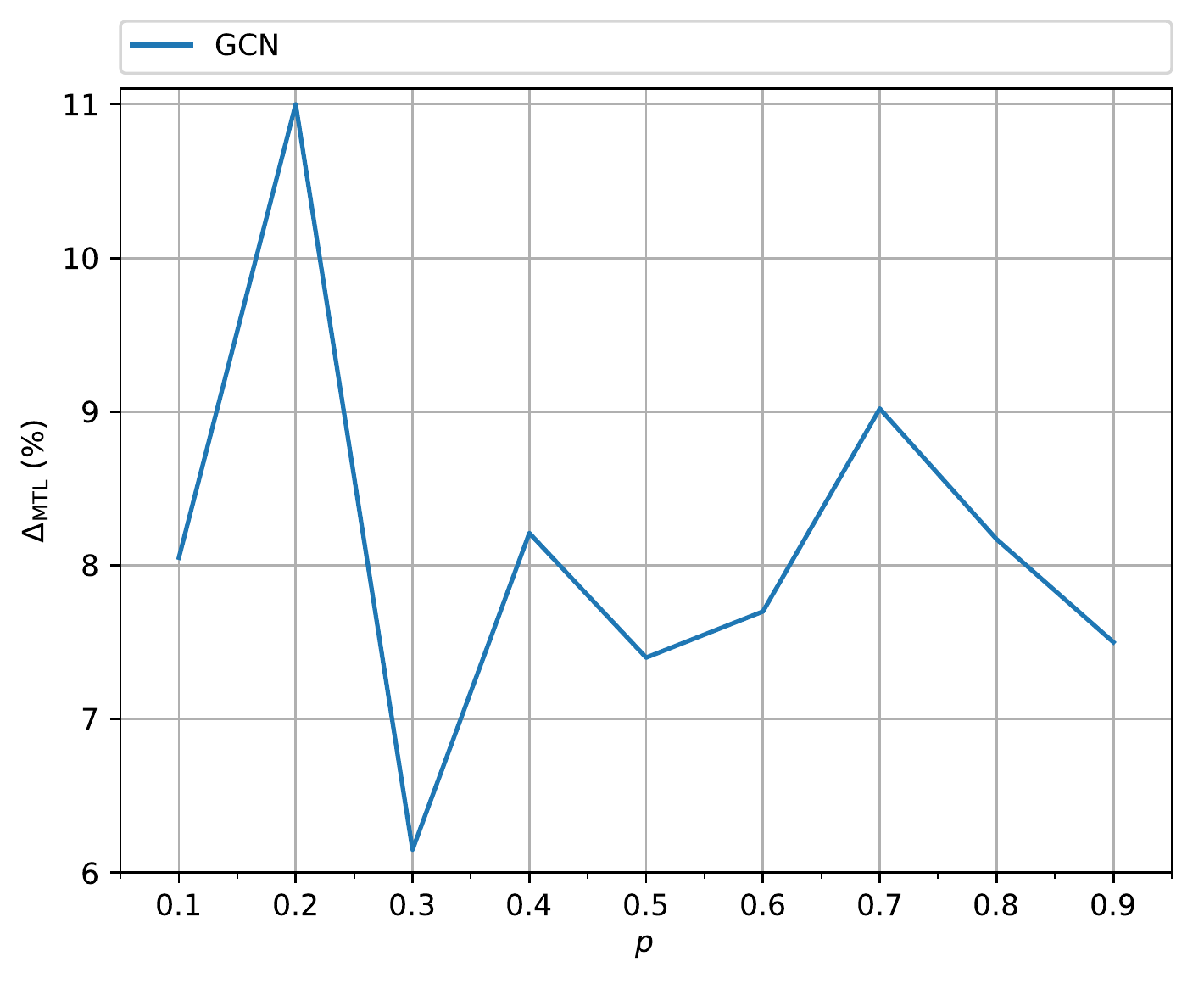}
  \captionof{figure}{Search over $p$.}
  \label{fig:SewerP}
\end{figure}

\begin{figure*}
\label{fig:adjmats}
\begin{subfigure}[b]{0.49\linewidth}
    \centering
    \includegraphics[width=\linewidth]{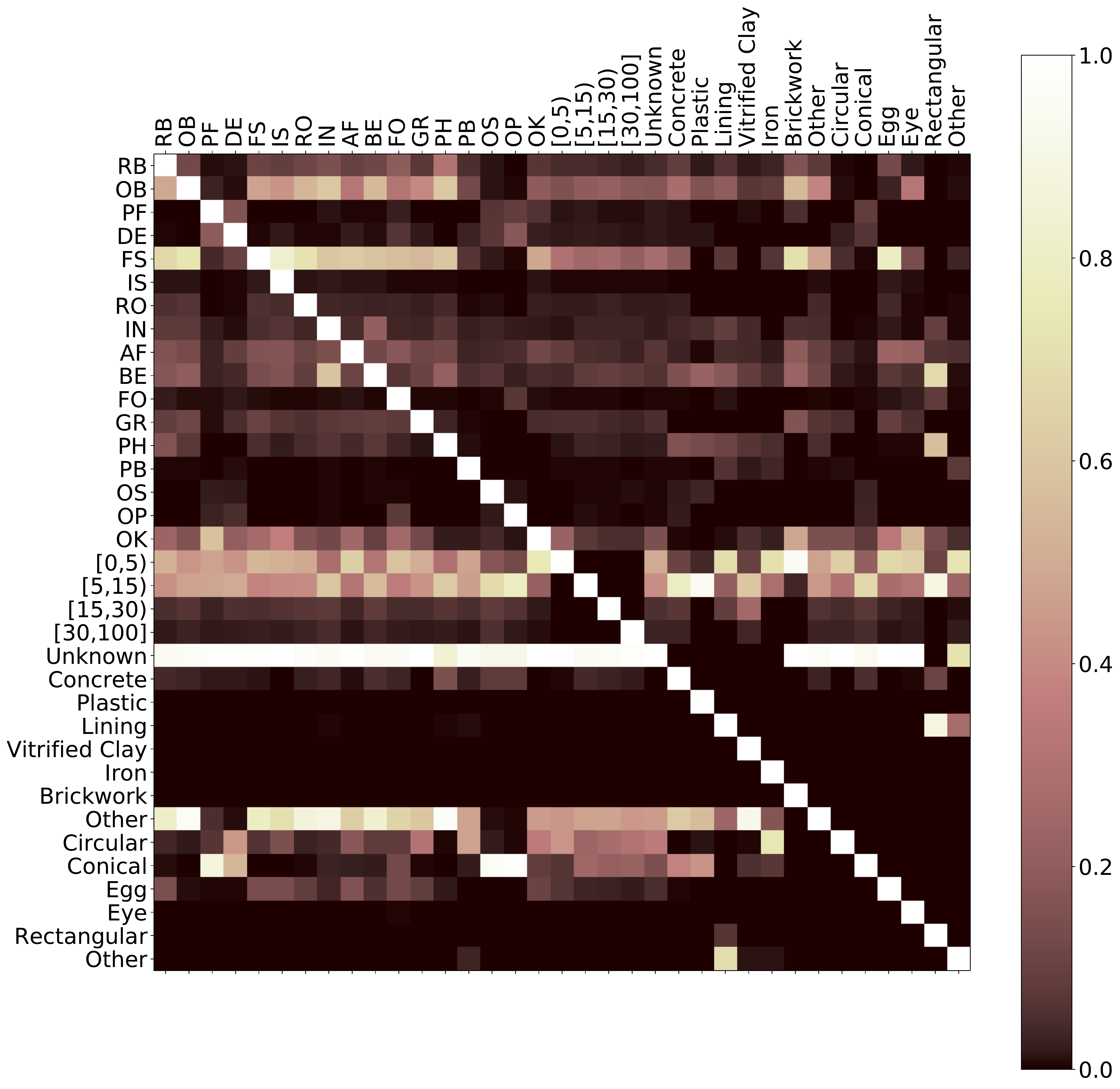}
    \caption{The conditional probability matrix based on the training labels.}
    \label{fig:condProb}
\end{subfigure}
\begin{subfigure}[b]{0.49\linewidth}
    \centering
    \includegraphics[width=\linewidth]{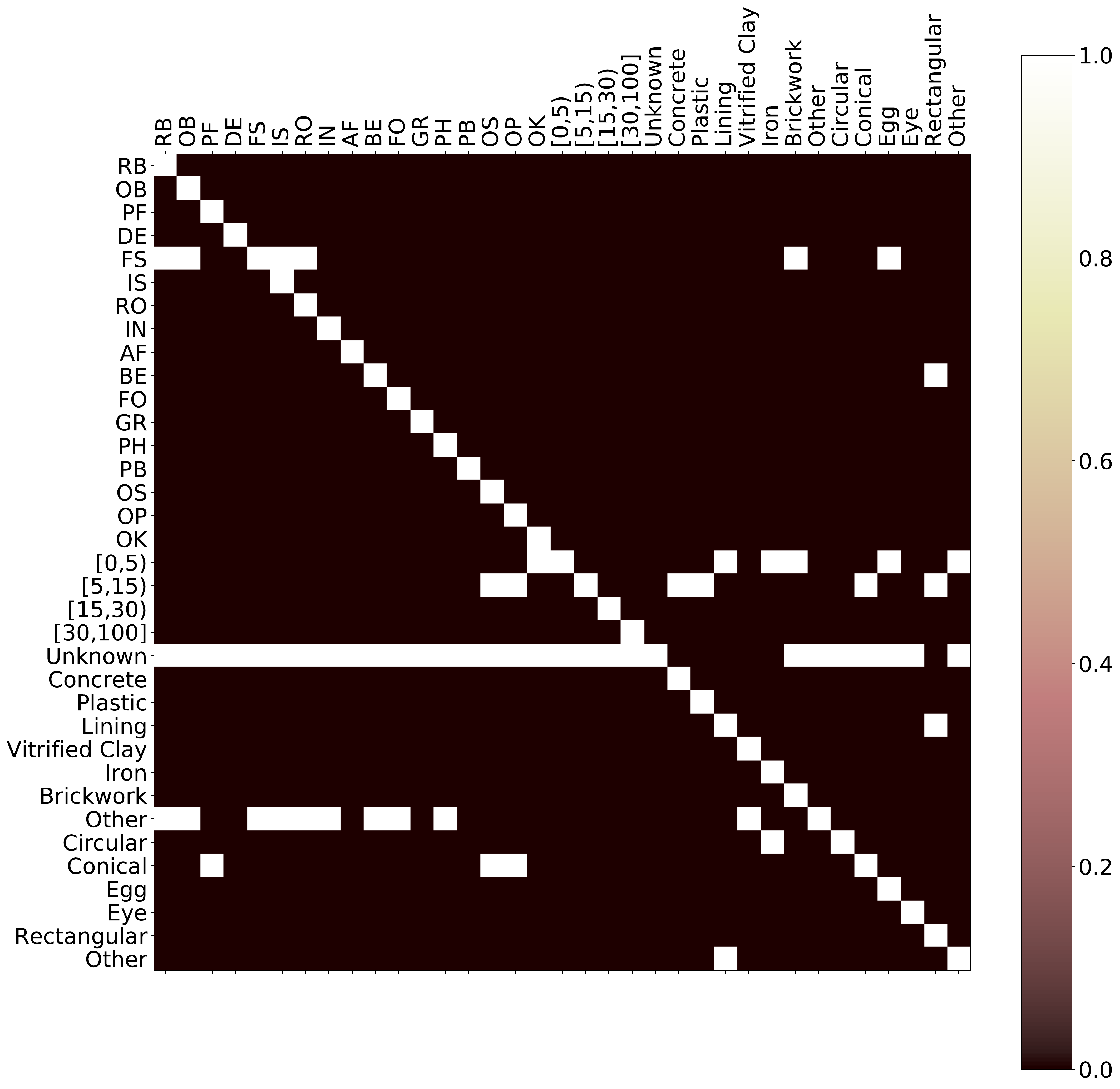}
    \caption{The re-weighted adjacency matrix obtained when $\tau=0.65$.}
    \label{fig:adGAT}
\end{subfigure}
\begin{subfigure}[b]{0.49\linewidth}
    \centering
    \includegraphics[width=\linewidth]{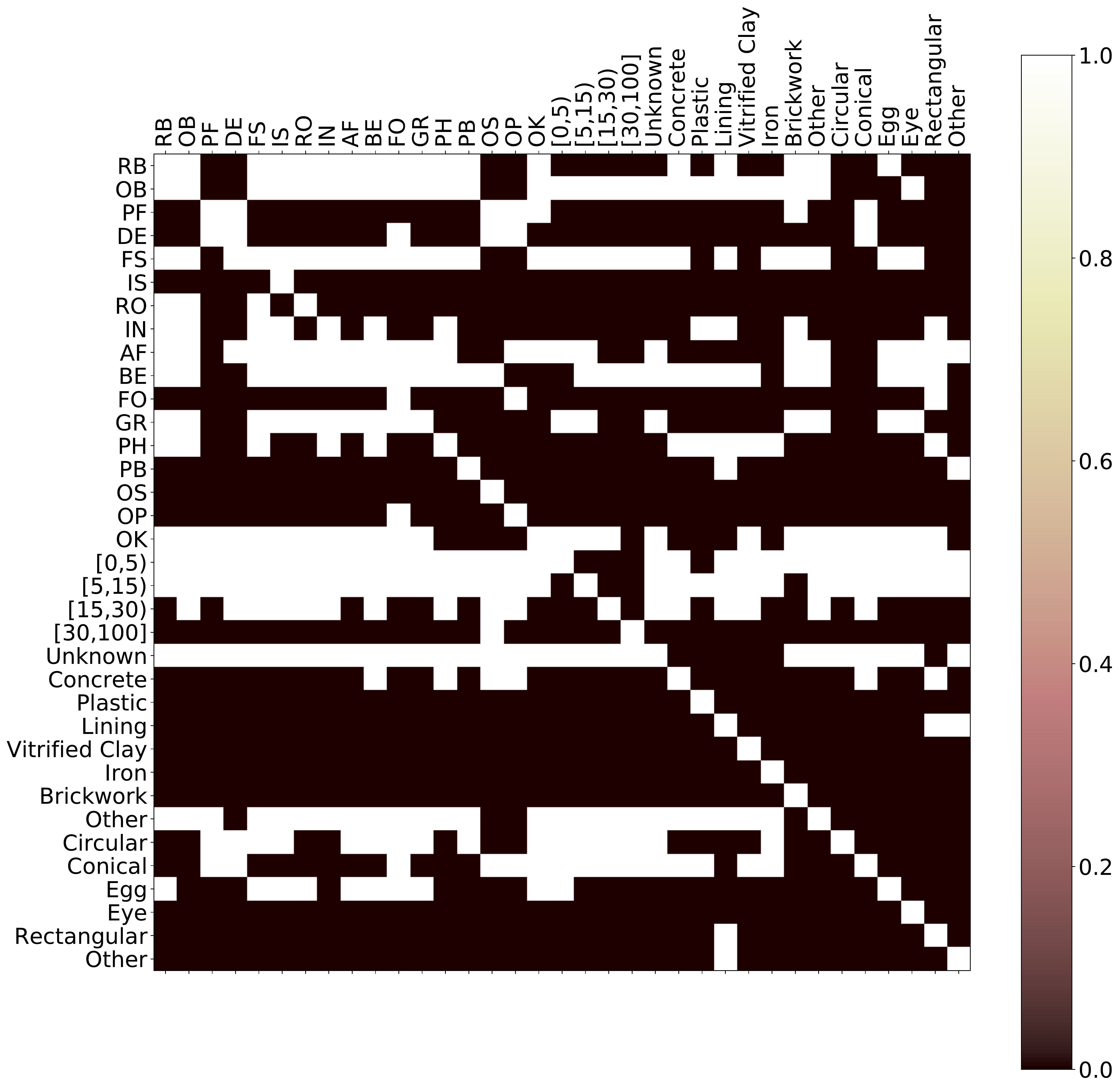}
    \caption{The re-weighted adjacency matrix obtained when $\tau=0.05$.}
    \label{fig:adj05b}
\end{subfigure}
\begin{subfigure}[b]{0.49\linewidth}
    \centering
    \includegraphics[width=\linewidth]{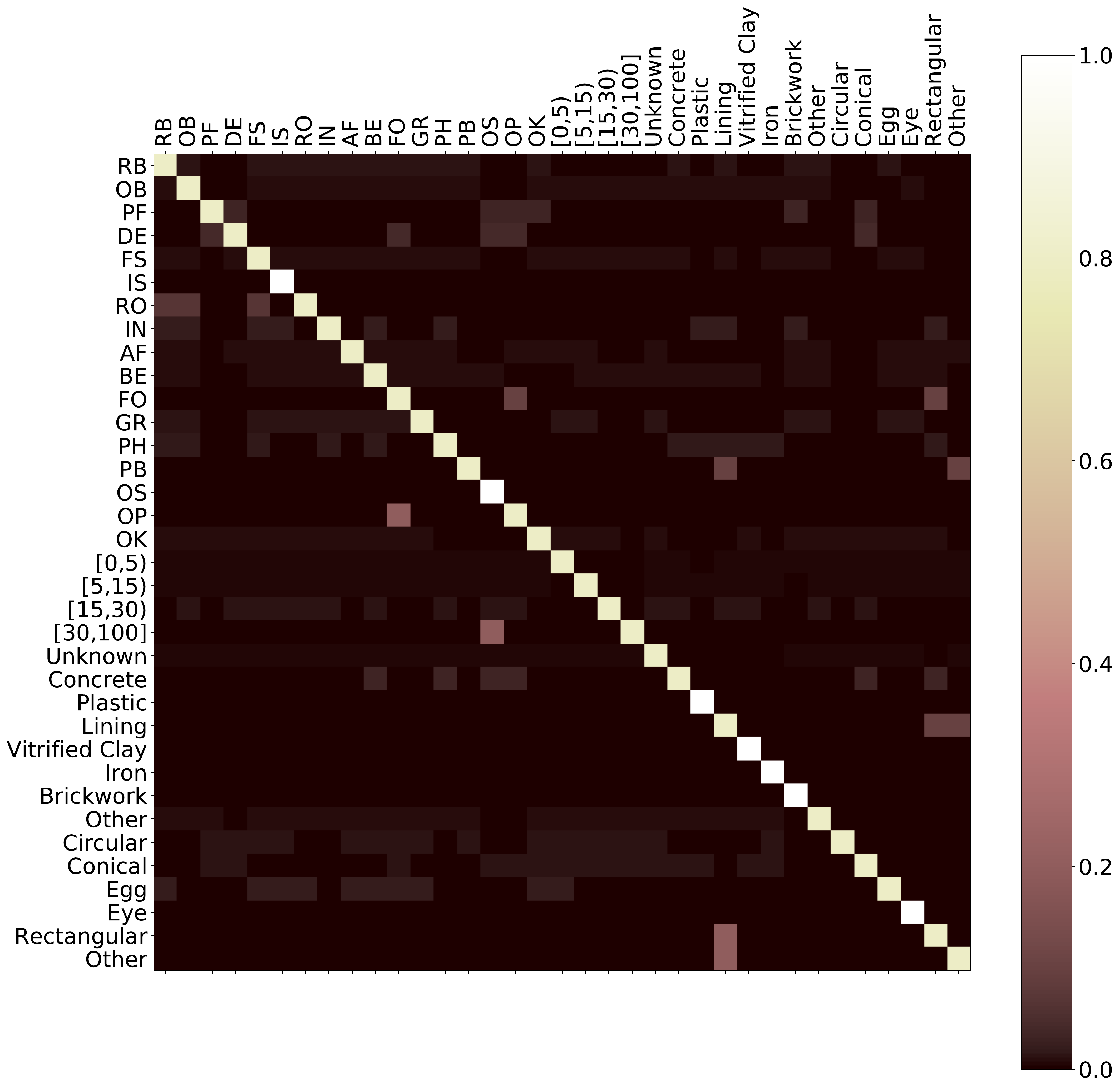}
    \caption{The re-weighted adjacency matrix when $\tau=0.05$ and $p=0.2$.}
    \label{fig:adjo5r}
\end{subfigure}
\caption{\textbf{Adjacency matrix construction.} We show the conditional probability matrix across task classes, as well as the constructed binary and reweighted adjacency matrices.}
\end{figure*}

The results of the sequential hyperparameter search on the Sewer-ML dataset are shown in Figures~\ref{fig:SewerGCNHeat}-\ref{fig:SewerP}. From these results we can conclude that the performance when using the GAT leads to more stable performances as the $\Delta_{\text{MTL}}$ in general does not vary as wildly as when using the GCN. However, when using the GCN we achieve in general higher $\Delta_{\text{MTL}}$. We can also observe that the adjacency matrix threshold $\tau$ has a large effect on the performance. Specifically, it is observable that using a low $\tau$ of 0.05 leads to good performance, which is only matched when $\tau$ is set to 0.65 and above for the GAT and 0.35 and above for the GCN.  Lastly, we observe that an increased neighbor node reweighting parameter $p$ leads to degraded performance, indicating that the center-node information is crucial. The conditional probability matrix, the binary matrices with $\tau$ set to 0.05 and 0.65, as well as the reweighted adjacency matrix with $\tau=0.05$ and $p = 0.2$ are shown in Figure~\ref{fig:condProb}-\ref{fig:adjo5r}.



\section{Optimization-Based MTL - In-Depth Results}\label{sec:opt}
We present the full results for the optimization-based method Dynamic Weight Averaging (DWA) \cite{MTAN2019} and Uncertainty estimation (Uncrt.) \cite{Uncertainty2018, Uncertainty22018}, see Table~\ref{tab:optRes}. The DWA task weighting method is initialized with $\lambda_\text{defect} = 0.90$, while Uncrt. is initialized with unit variances for each task.
From the results we observe that the DWA method performs worse than the STL networks on nearly every task. The Uncrt. method improves the shape and material MF1 compared to the STL networks, but suffers from poor defect classification rate.

\section{Effect of $\lambda_\text{defect}$ - In-Depth Results}\label{sec:lambda}
We show the in-depth results for each tested setting of $\lambda_\text{defect}$ on the validation split for the R50-MTL baseline as well as CT-GNNs, see Table~\ref{tab:sewerLambda}. We observe that a larger $\lambda_\text{defect}$ leads to a higher $\Delta_\text{MTL}$ due to a higher $\text{F}2_\text{CIW}$. However, it also leads to a lower material MF1 score, as we observe that the material MF1 score peaks at 90.5\% for the CT-GNNs when $\lambda_\text{defect}=0.50$, and decreases to 82-86\% when $\lambda_\text{defect}=0.90$. 

\section{Combining the MTAN Encoder and CT-GNN Decoder - In-Depth Results}\label{sec:mtan}
We present the in-depth results of the ablation studies investigating the combination of MTAN encoder and the CT-GNN Decoder, see Table~\ref{tab:sewerMTAN}. The methods were only evaluated on the validation split. From the results we see that the $\Delta_\text{MTL}$ is increased by introducing the CT-GNN, and that the combination with the CT-GCN outperforms using a hard-shared encoder. We observe that the noticeable difference is in the defect classification task where the performance is increased by 0.6-0.7 percentage points on the $\text{F}2_\text{CIW}$ metric. 

\section{CT-GNN Success and Failure Cases}\label{sec:success}
\begin{figure*}[!t]
\centering
\begin{subfigure}[b]{0.3\linewidth}
  \centering
  \includegraphics[width=0.98\linewidth]{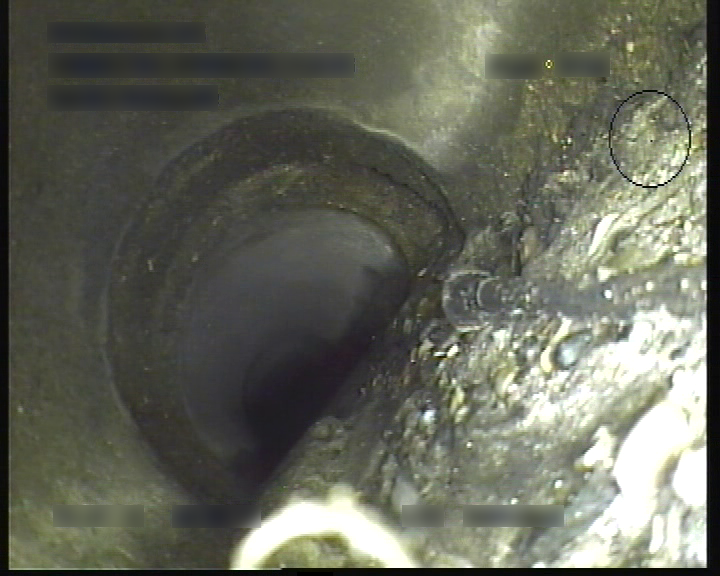}
  \qquad
    \resizebox{0.98\linewidth}{!}{%
    \begin{tabular}[b]{ccc}\toprule
      Task &  Ground Truth   & CT-GNN \\ \cmidrule{1-3}
       Defect & RB,OB,FS,AF  & RB,OB,FS,AF\\
       Water & [0\%, 5\%) & [0\%, 5\%) \\
       Shape & Circular  & Circular \\
       Material & Concrete & Concrete\\ \bottomrule
    \end{tabular}%
    }
 \end{subfigure}%
\begin{subfigure}[b]{0.3\linewidth}
  \centering
  \includegraphics[width=0.98\linewidth]{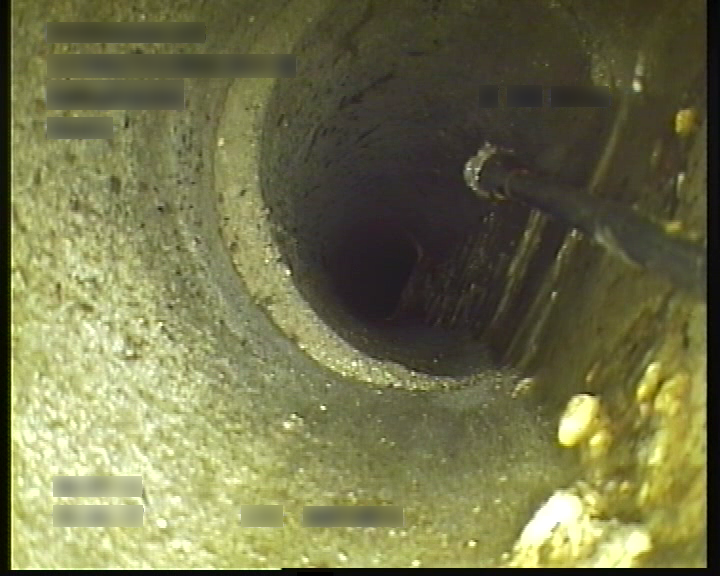}
  \qquad
    \resizebox{0.9\linewidth}{!}{%
    \begin{tabular}[b]{ccc}\toprule
      Task & Ground Truth & CT-GNN \\ \cmidrule{1-3}
       Defect & FS,AF& FS,AF\\
       Water & [5\%, 15\%) & [5\%, 15\%) \\
       Shape & Circular  & Circular \\
       Material & Concrete & Concrete\\ \bottomrule
    \end{tabular}%
    }
 \end{subfigure}%
\begin{subfigure}[b]{0.3\linewidth}
  \centering
  \includegraphics[width=0.98\linewidth]{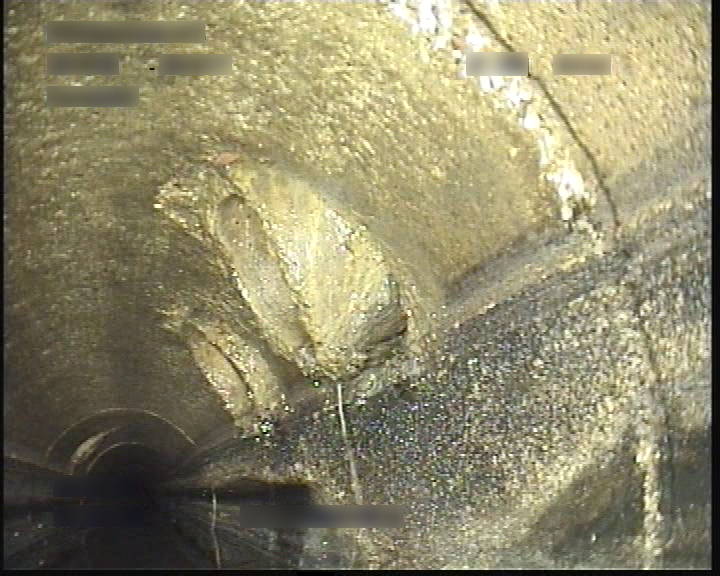}
  \qquad
    \resizebox{0.98\linewidth}{!}{%
    \begin{tabular}[b]{ccc}\toprule
      Task & Ground Truth & CT-GNN \\ \cmidrule{1-3}
       Defect & FS,PH  & FS,PH\\
       Water & [30\%, 100\%] & [30\%, 100\%] \\
       Shape & Circular  & Circular \\
       Material & Concrete & Concrete\\ \bottomrule
    \end{tabular}%
    }
 \end{subfigure}
 \vspace{2mm}
\begin{subfigure}[b]{0.3\linewidth}
  \centering
  \includegraphics[width=0.98\linewidth]{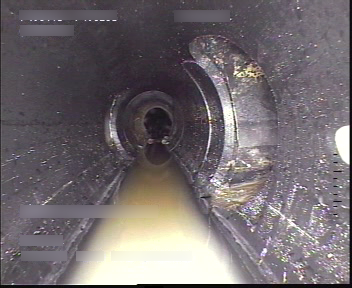}
  \qquad
    \resizebox{0.98\linewidth}{!}{%
    \begin{tabular}[b]{ccc}\toprule
      Task & Ground Truth & CT-GNN \\ \cmidrule{1-3}
       Defect & RB,PB  & RB,PB\\
       Water & [5\%, 15\%) & [5\%, 15\%) \\
       Shape & Circular  & Circular \\
       Material & Plastic & Plastic\\ \bottomrule
    \end{tabular}%
    }
 \end{subfigure}%
\begin{subfigure}[b]{0.3\linewidth}
  \centering
  \includegraphics[width=0.98\linewidth]{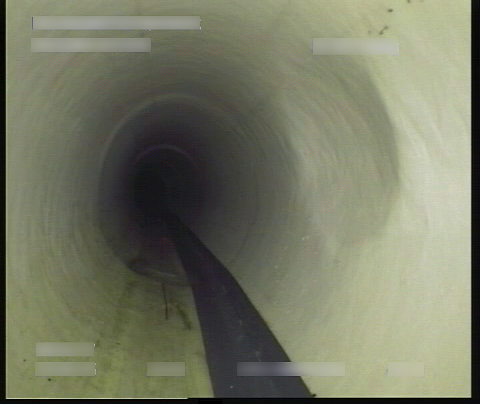}
  \qquad
    \resizebox{0.98\linewidth}{!}{%
    \begin{tabular}[b]{ccc}\toprule
      Task & Ground Truth & CT-GNN \\ \cmidrule{1-3}
       Defect & DE  & DE\\
       Water & [5\%, 15\%) & [5\%, 15\%) \\
       Shape & Circular  & Circular \\
       Material & Lining & Lining\\ \bottomrule
    \end{tabular}%
    }
 \end{subfigure}%
\begin{subfigure}[b]{0.3\linewidth}
  \centering
  \includegraphics[width=0.98\linewidth]{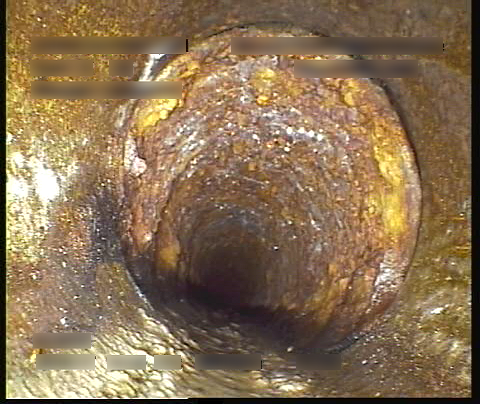}
  \qquad
    \resizebox{0.93\linewidth}{!}{%
    \begin{tabular}[b]{ccc}\toprule
      Task & Ground Truth & CT-GNN \\ \cmidrule{1-3}
       Defect & OB,OK  & OB,OK\\
       Water & [0\%, 5\%) & [0\%, 5\%) \\
       Shape & Circular  & Circular \\
       Material & Iron & Iron\\ \bottomrule
    \end{tabular}%
    }
 \end{subfigure}
 
  \caption{\textbf{Examples of correct classifications with the CT-GNN.} Example cases where the CT-GNN correctly classifies all four tasks.}
  \label{fig:success}
\end{figure*}

\begin{figure*}[!t]
\centering
\begin{subfigure}[b]{0.3\linewidth}
  \centering
  \includegraphics[width=0.98\linewidth]{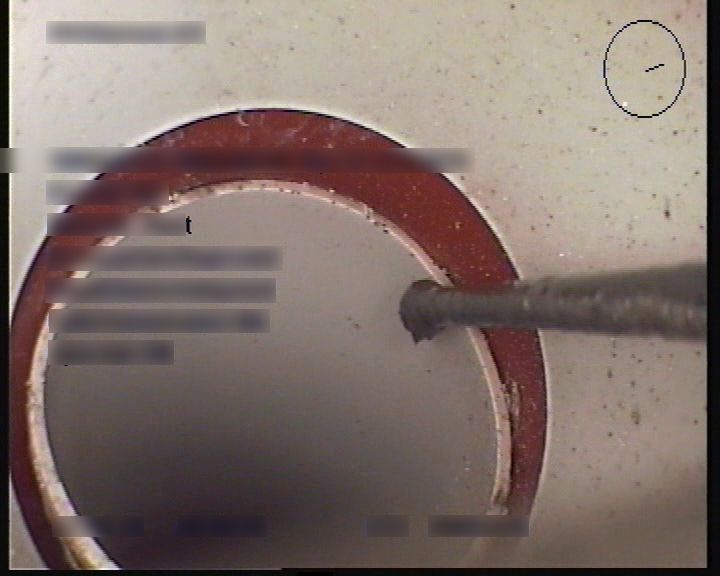}
  \qquad
    \resizebox{0.98\linewidth}{!}{%
    \begin{tabular}[b]{ccc}\toprule
      Task &  Ground Truth   & CT-GNN \\ \cmidrule{1-3}
       Defect & OK  & OK,{\color{red}FS}\\
       Water & [0\%, 5\%) & [0\%, 5\%) \\
       Shape & Circular  & Circular \\
       Material & Plastic & Plastic\\ \bottomrule
    \end{tabular}%
    }
 \end{subfigure}%
\begin{subfigure}[b]{0.3\linewidth}
  \centering
  \includegraphics[width=0.98\linewidth]{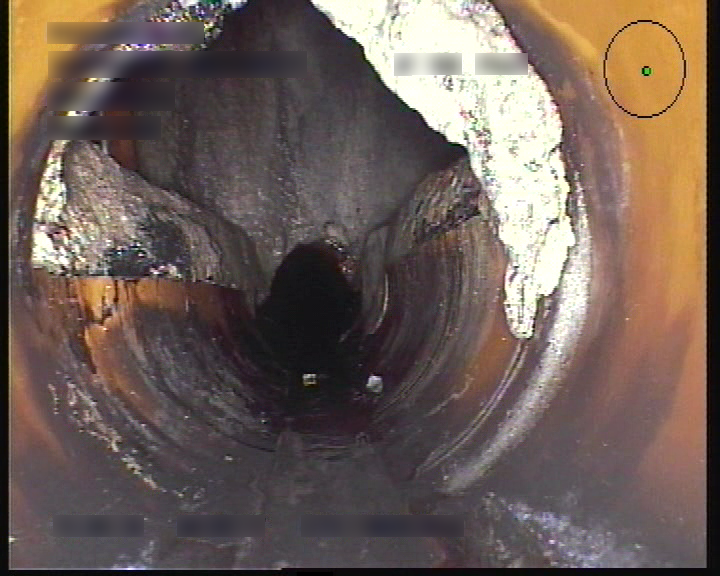}
  \qquad
    \resizebox{1\linewidth}{!}{%
    \begin{tabular}[b]{ccc}\toprule
      Task & Ground Truth & CT-GNN \\ \cmidrule{1-3}
       Defect & BE& BE,{\color{red}OK}\\
       Water & [0\%, 5\%) & {\color{red}[5\%, 15\%)} \\
       Shape & Circular  & Circular \\
       Material & Plastic & Plastic\\ \bottomrule
    \end{tabular}%
    }
 \end{subfigure}%
\begin{subfigure}[b]{0.3\linewidth}
  \centering
  \includegraphics[width=0.98\linewidth]{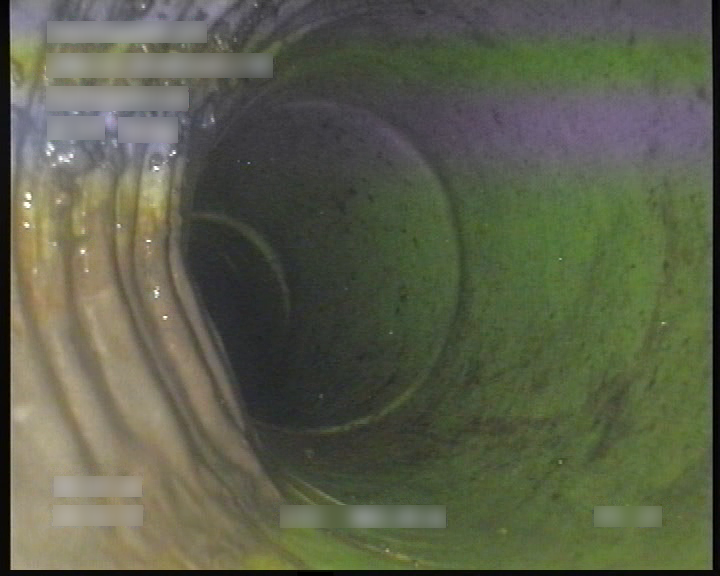}
  \qquad
    \resizebox{0.98\linewidth}{!}{%
    \begin{tabular}[b]{ccc}\toprule
      Task & Ground Truth & CT-GNN \\ \cmidrule{1-3}
       Defect & DE,OK  & {\color{red}{OB}},OK\\
       Water & [0\%, 5\%) & [0\%, 5\%) \\
       Shape & Circular  & Circular \\
       Material & Lining & Lining\\ \bottomrule
    \end{tabular}%
    }
 \end{subfigure}
 \vspace{2mm}
\begin{subfigure}[b]{0.3\linewidth}
  \centering
  \includegraphics[width=0.98\linewidth]{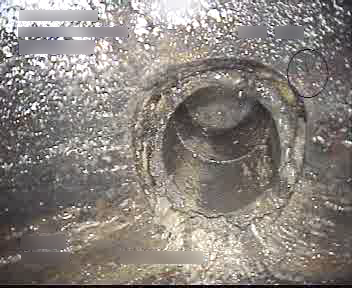}
  \qquad
    \resizebox{0.98\linewidth}{!}{%
    \begin{tabular}[b]{ccc}\toprule
      Task & Ground Truth & CT-GNN \\ \cmidrule{1-3}
       Defect & PF,OS  & {\color{red}OB,FS,PH}\\
       Water & [5\%, 15\%) & {\color{red}[30\%, 100\%]} \\
       Shape & Conical  & {\color{red}Circular} \\
       Material & Lining & {\color{red}Concrete}\\ \bottomrule
    \end{tabular}%
    }
 \end{subfigure}%
\begin{subfigure}[b]{0.3\linewidth}
  \centering
  \includegraphics[width=0.98\linewidth]{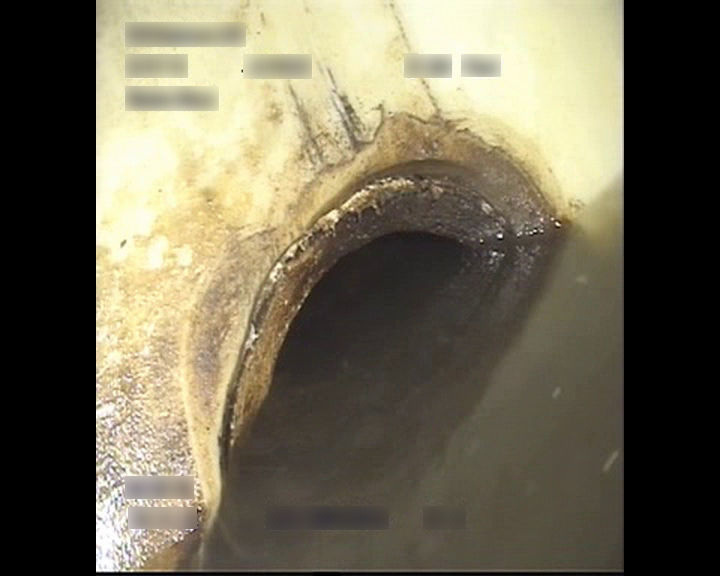}
  \qquad
    \resizebox{0.98\linewidth}{!}{%
    \begin{tabular}[b]{ccc}\toprule
      Task & Ground Truth & CT-GNN \\ \cmidrule{1-3}
       Defect & OS  & {\color{red}None}\\
       Water & [15\%, 30\%) & {\color{red}[30\%, 100\%]} \\
       Shape & Circular  & {\color{red}Conical} \\
       Material & Plastic & {\color{red}Lining}\\ \bottomrule
    \end{tabular}%
    }
 \end{subfigure}%
\begin{subfigure}[b]{0.3\linewidth}
  \centering
  \includegraphics[width=0.98\linewidth]{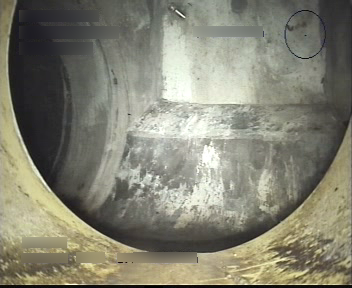}
  \qquad
    \resizebox{0.9\linewidth}{!}{%
    \begin{tabular}[b]{ccc}\toprule
      Task & Ground Truth & CT-GNN \\ \cmidrule{1-3}
       Defect & OK  & {\color{red}None}\\
       Water & [5\%, 15\%) & {\color{red}[0\%, 5\%)} \\
       Shape & Circular  & {\color{red}Conical} \\
       Material & Plastic & {\color{red}Lining}\\ \bottomrule
    \end{tabular}%
    }
 \end{subfigure}
 
  \caption{\textbf{Examples of incorrect classifications with the CT-GNN.} Example cases where the CT-GNN incorrectly classifies some or all four tasks. Incorrect classifications are denoted in {\color{red}red}.}
  \label{fig:fails}
\end{figure*}

We show several cases where the CT-GNN decoder correctly classifies all tasks, shown in Figure~\ref{fig:success}, as well as cases where some or all tasks are misclassified, shown in Figure~\ref{fig:fails}.

We observe that the the CT-GNN performs well when several defects occur at the same time at different distances to the camera (see top left example), as well as subtle defects such as the distortion in the bottom middle example and crack in the bottom left example. Similarly, this can be observed in the top right example where the high water level is detected even though it is partially occluded and unlit.  Lastly it can correctly handle rare classes such as the iron material in the bottom right example.

In Figure~\ref{fig:fails} we observe that the the CT-GNN misclassify irregularities in the pipe geometry as displaced pipes (FS) or construction changes (OK), as seen in the top right and top middle examples. In both cases the predictions is understandable as the internal reparation is shifted (top left) and the camera is placed right before a well (top middle). In the top right case the deformation is observed as a surface damage, which is understandable due to the folds of the deformation. For the cases where all classifications are incorrect, we see that the CT-GNN decoder misclassifies several tasks due to limited context introduced by the camera perspective.

\clearpage

\begin{table*}[!t]
\centering
\caption{\textbf{Effect of optimization-based methods.} In-depth results for two optimization-based methods, DWA \cite{MTAN2019} and the uncertainty (Uncrt.) based method \cite{Uncertainty2018, Uncertainty22018}. TW indicates the task weighting method used and \#P indicates the number of parameters in millions. The best performance in each column is denoted in \textbf{bold}.}
\label{tab:optRes}
\begin{tabular}{llccrrccccccccc} \toprule
&\multicolumn{3}{c}{Model} & \multicolumn{1}{c}{Overall} & \multicolumn{2}{c}{Defect} & \multicolumn{2}{c}{Water} & \multicolumn{2}{c}{Shape} & \multicolumn{2}{c}{Material}  \\ 
\cmidrule(r){2-4} \cmidrule(r){5-5} \cmidrule(r){6-7} \cmidrule(r){8-9} \cmidrule(r){10-11} \cmidrule(r){12-13} 

& Encoder & TW & \#P  & \textbf{$\Delta_{\text{MTL}}$}& \textbf{$\text{F}2_{\text{CIW}}$} & \textbf{$\text{F}1_{\text{Normal}}$} & \textbf{MF1} & \textbf{mF1} & \textbf{MF1} & \textbf{mF1} & \textbf{MF1} & \textbf{mF1} \\ \cmidrule(r){1-13} 

\parbox[t]{2mm}{\multirow{3}{*}{\rotatebox[origin=c]{90}{Val.}}} 

& STL & - &94.0& +\textbf{0.00} & \textbf{58.42} & \textbf{92.42} & \textbf{69.11} & \textbf{79.71} & 46.55 & 98.06 & 65.99 & \textbf{96.71}   \\ 
& R50-MTL  &   DWA & 23.5 & -15.70 & 34.22 &	86.57&	53.43&	70.83&	37.68&	98.18&	53.50&	90.79 \\
& R50-MTL & Uncrt. & 23.5& -4.07 & 24.80 &	86.80 &	62.00 &	75.31 &	\textbf{67.30} &	\textbf{99.19} &	\textbf{67.46} &	95.66  \\
\cmidrule(r){1-13}\morecmidrules\cmidrule(r){1-13}

\parbox[t]{2mm}{\multirow{3}{*}{\rotatebox[origin=c]{90}{Test}}} 
&STL & - &94.0 & +\textbf{0.00} & \textbf{57.48} & \textbf{92.16} & \textbf{69.87} & \textbf{80.09} & 56.15 & 97.59 & 69.02 & \textbf{96.67}   \\ 
& R50-MTL &   DWA & 23.5 &  -11.57 &34.84&	86.20&	54.30&	71.03&	59.27&	97.81&	60.39&	90.49 \\
& R50-MTL  & Uncrt. & 23.5& -3.78 & 26.30 &	86.48 &	63.01 &	76.15 &	\textbf{79.69} &	\textbf{98.99} &	\textbf{70.84} &	95.59    \\ \bottomrule

\end{tabular}
\end{table*}

\begin{table*}[!t]
\centering
\caption{\textbf{Effect of $\lambda_\text{defect}$.} We compare the performance of the R50-MTL baseline and CT-GNN heads when training with different $\lambda_\text{defect}$ values. Evaluated on the validation split. The best performance in each column is denoted in \textbf{bold} per method.}
\label{tab:sewerLambda}
\begin{tabular}{clcccccccrc} \toprule
\multicolumn{2}{c}{Model}& \multicolumn{1}{c}{Overall}  & \multicolumn{2}{c}{Defect} & \multicolumn{2}{c}{Water} & \multicolumn{2}{c}{Shape} & \multicolumn{2}{c}{Material}  \\ 
\cmidrule(r){1-2} \cmidrule(r){3-3} \cmidrule(r){4-5} \cmidrule(r){6-7} \cmidrule(r){8-9} \cmidrule(r){10-11} 

Model & $\lambda_\text{defect}$ & \textbf{$\Delta_{\text{MTL}}$}& \textbf{$\text{F}2_{\text{CIW}}$} & \textbf{$\text{F}1_{\text{Normal}}$} & \textbf{MF1} & \textbf{mF1} & \textbf{MF1} & \textbf{mF1} & \textbf{MF1} & \textbf{mF1}   \\ \cmidrule(r){1-11} 

\parbox[t]{2mm}{\multirow{7}{*}{\rotatebox[origin=c]{90}{R50-MTL}}}
& 0.25  & +5.45 &32.86&	88.40&	69.42&	79.85&	74.72&	99.21&	84.64&	97.83  \\
& 0.33 & +6.22 & 39.85&	89.08&	69.18&	79.89&	70.29&	99.31&	86.61&	97.96  \\
& 0.50  & +6.91 & 40.78&	89.31&	69.35&	79.90&	71.58&	99.25&	87.21&	97.75  \\
& 0.67  & +\textbf{11.11} & 52.53&	90.69&	70.19&	80.22&	\textbf{75.74}&	99.38&	\textbf{87.83}&	98.11  \\
& 0.75 & +9.99 & 56.31&	91.41&	70.15&	80.42&	69.91&	\textbf{99.40}&	85.13&	\textbf{98.30}  \\
& 0.90 & +10.36 & 59.73&	\textbf{91.87}&	\textbf{70.51}&	\textbf{80.47}&	71.64&	99.34&	80.28&	98.09  \\
& 0.95 & +10.40 & \textbf{60.34}&	91.85&	69.35&	80.02&	71.85&	99.19&	81.26&	97.82  \\ \cmidrule{1-11}

\parbox[t]{2mm}{\multirow{7}{*}{\rotatebox[origin=c]{90}{CT-GCN}}}
& 0.25  & +5.79   & 39.44&	88.77&	69.76&	79.63&	73.67&	99.27&	80.06&	97.79  \\
& 0.33 & +7.45    & 42.56&	89.12&	69.36&	79.82&	72.20&	99.20&	87.40&	97.81  \\
& 0.50  & +11.00  & 50.35&	90.01&	70.04&	79.98&	75.80&	\textbf{99.44}&	\textbf{90.54}&	97.96  \\
& 0.67  & +10.20  & 54.67&	90.64&	69.78&	79.92&	72.94&	99.35&	85.31&	98.06  \\
& 0.75 & +10.75   & 57.71&	91.11&	70.48&	80.21&	70.95&	99.37&	86.27&	\textbf{98.21}  \\
& 0.90 & +\textbf{12.39 }  & 61.35&	91.84&	\textbf{70.57}&	\textbf{80.47}&	\textbf{76.17}&	99.33&	82.63&	98.18  \\
& 0.95 & +9.05    & \textbf{62.10}&	\textbf{92.01}&	69.95&	80.04&	67.36&	99.11&	77.83&	97.89  \\ \cmidrule{1-11}

\parbox[t]{2mm}{\multirow{7}{*}{\rotatebox[origin=c]{90}{CT-GAT}}}
& 0.25  & +7.69  & 37.02&	88.69&	70.06&	80.18&	75.47&	99.45&	89.40&	97.89  \\
& 0.33 & +5.70   & 42.17&	89.09&	69.54&	79.96&	71.72&	99.37&	78.80&	97.95  \\
& 0.50  & +10.33 & 49.96&	89.98&	69.69&	79.90&	73.90&	99.41&	\textbf{90.52}&	98.06  \\
& 0.67  & +10.20 & 55.26&	90.69&	69.80&	80.38&	72.41&	99.40&	84.90&	98.12  \\
& 0.75 & +12.10  & 58.37&	91.45&	70.46&	\textbf{80.43}&	\textbf{76.82}&	\textbf{99.46}&	83.75&	\textbf{98.35}  \\
& 0.90 & +\textbf{12.81}  & \textbf{61.70}&	\textbf{91.94}&	\textbf{70.57}&	\textbf{80.43}&	74.53&	99.40&	86.63&	98.24  \\
& 0.95 & +10.65  & 60.95&	92.03&	69.01&	79.59&	70.75&	99.18&	83.99&	97.84  \\

\bottomrule
\end{tabular}
\end{table*}

\begin{table*}[!t]
\centering
\caption{\textbf{Effect of encoder.} We compare the effect of training CT-GNN using GCN and GAT with the MTAN encoder, and with fixed task weights. \#P indicates the number of parameters in millions. Evaluated on the validation split. The best performance in each column is denoted in \textbf{bold}.}
\label{tab:sewerMTAN}
\begin{tabular}{llccccccccrc} \toprule
\multicolumn{3}{c}{Model}& \multicolumn{1}{c}{Overall}  & \multicolumn{2}{c}{Defect} & \multicolumn{2}{c}{Water} & \multicolumn{2}{c}{Shape} & \multicolumn{2}{c}{Material}  \\ 
\cmidrule(r){1-3} \cmidrule(r){4-4} \cmidrule(r){5-6} \cmidrule(r){7-8} \cmidrule(r){9-10} \cmidrule(r){11-12} 

Encoder & CT-GNN & \#P& \textbf{$\Delta_{\text{MTL}}$}& \textbf{$\text{F}2_{\text{CIW}}$} & \textbf{$\text{F}1_{\text{Normal}}$} & \textbf{MF1} & \textbf{mF1} & \textbf{MF1} & \textbf{mF1} & \textbf{MF1} & \textbf{mF1}   \\ \cmidrule(r){1-12} 

MTAN & \xmark & 48.2 & +10.40 &61.21&	\textbf{92.10}&	70.06&	\textbf{80.59}&	68.34&	99.40&	83.48&	98.25  \\
MTAN & GCN  &  49.9  & +\textbf{12.72} & 61.86&	91.99&	\textbf{71.39}&	80.53&	\textbf{75.42}&	\textbf{99.46}&	\textbf{83.77}&	98.25    \\ 
MTAN & GAT  & 48.6& +11.48 & \textbf{61.92}&	92.03&	70.95&	80.50&	71.17&	99.39&	83.65&	\textbf{98.29 }    \\ 

\bottomrule
\end{tabular}
\end{table*}

\clearpage
\onecolumn
\def\tableTemp{\thetable}
\renewcommand\tablename{Figure}
\setcounter{table}{\thefigure}
\begin{longtable}[c]{m{1.5cm}m{2.69cm}m{2.69cm}m{2.69cm}m{2.69cm}m{2.69cm}}
\caption{\textbf{Water level class examples.} Example images of the four considered water level classes.}\\
\endfirsthead
\multicolumn{6}{c}%
        {{ Figure \thetable: \textbf{Continued from previous page}}} \\
\endhead
\multicolumn{6}{r}{{Continued on next page}} 
\endfoot
\endlastfoot
\centering
\textbf{$[0\%,5\%)$}     & \includegraphics[width=1\linewidth]{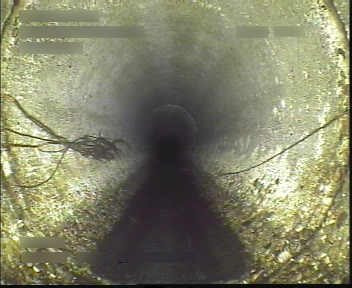} & \includegraphics[width=1\linewidth]{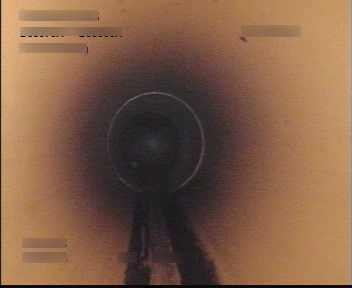} & \includegraphics[width=1\linewidth]{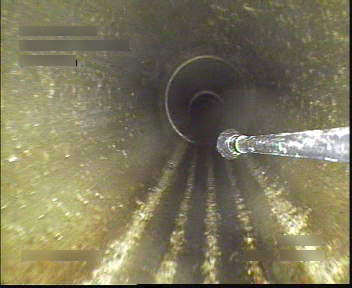} & \includegraphics[width=1\linewidth]{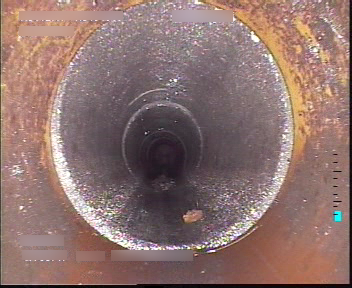} & \includegraphics[width=1\linewidth]{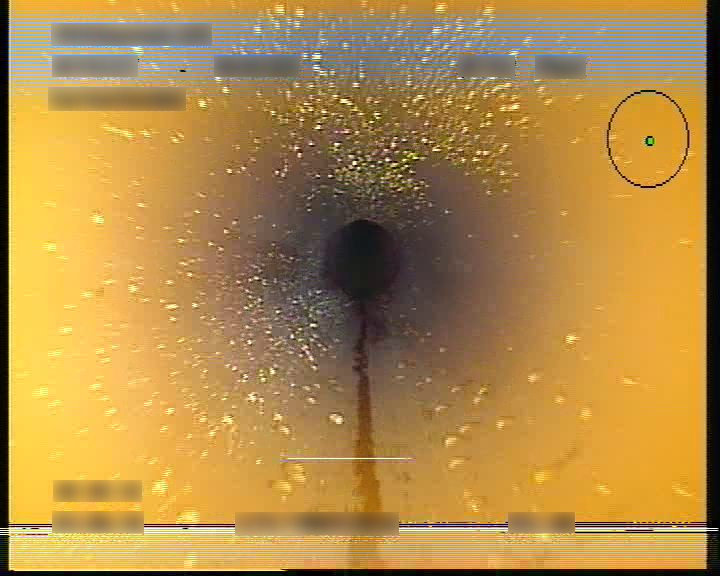}  \\
\textbf{$[5\%, 15\%)$}     & \includegraphics[width=1\linewidth]{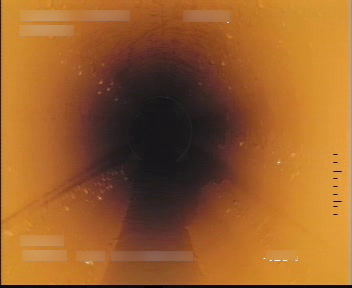} & \includegraphics[width=1\linewidth]{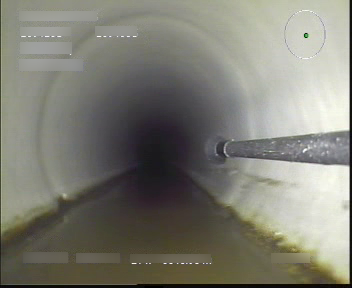} & \includegraphics[width=1\linewidth]{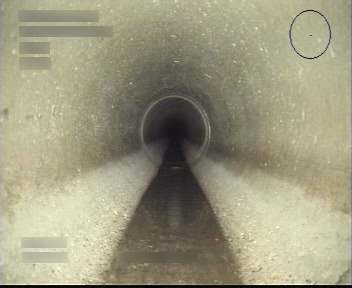} & \includegraphics[width=1\linewidth]{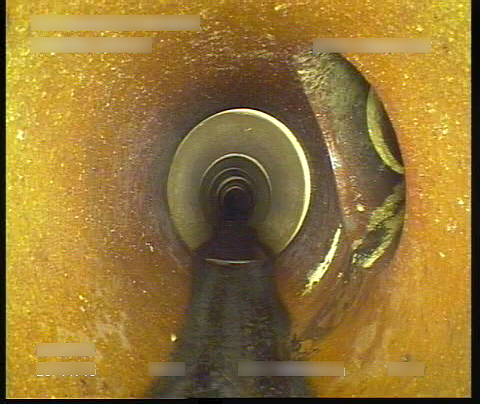} & \includegraphics[width=1\linewidth]{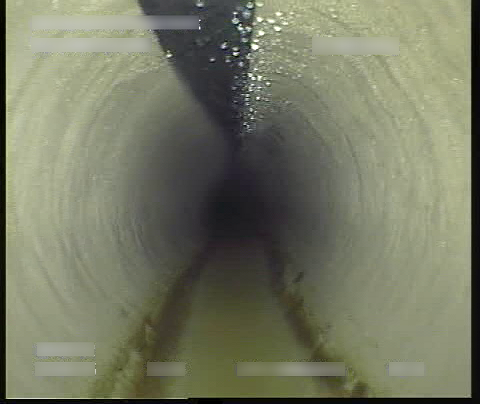}  \\
\textbf{$[15\%, 30\%)$}     & \includegraphics[width=1\linewidth]{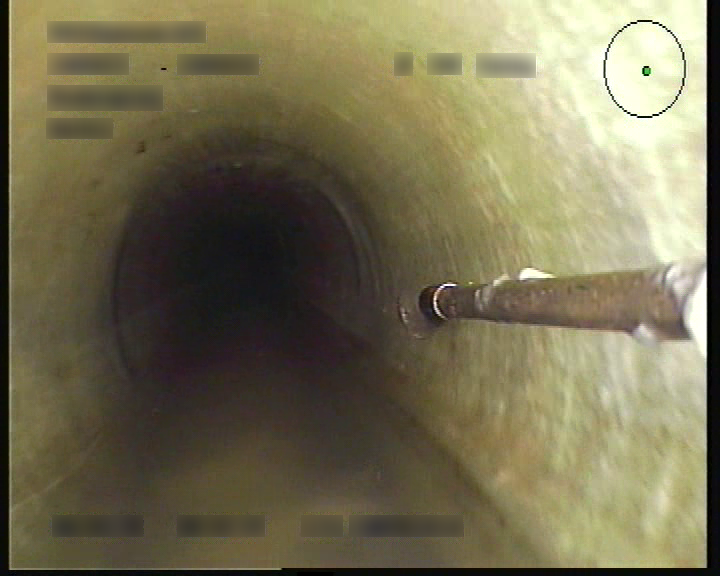} & \includegraphics[width=1\linewidth]{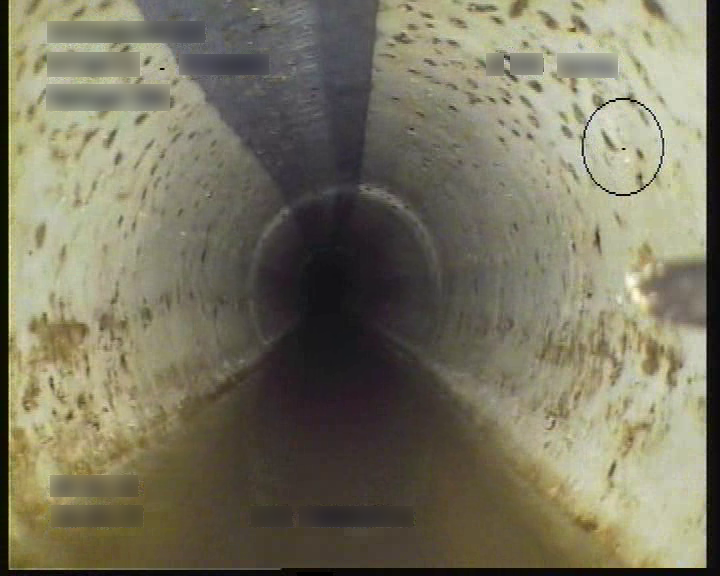} & \includegraphics[width=1\linewidth]{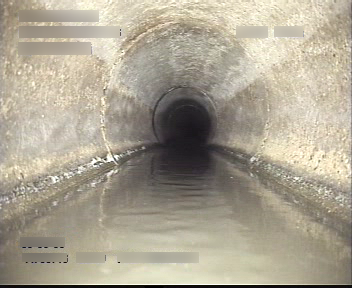} & \includegraphics[width=1\linewidth]{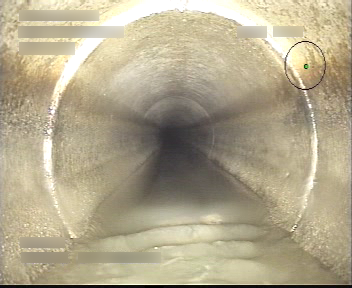} & \includegraphics[width=1\linewidth]{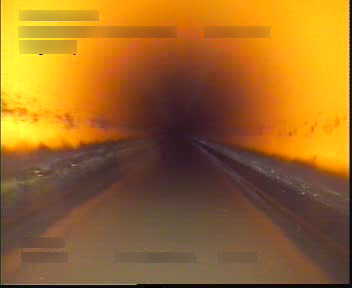}  \\
\textbf{$[30\%, 100\%]$}     & \includegraphics[width=1\linewidth]{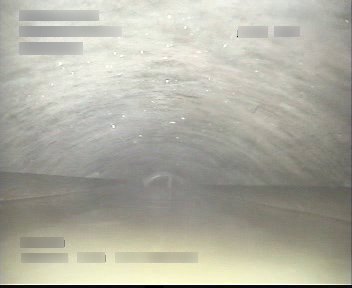} & \includegraphics[width=1\linewidth]{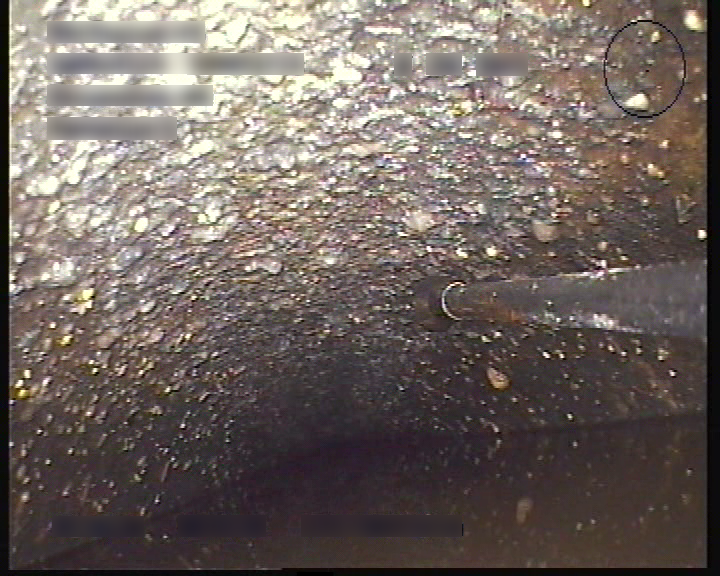} & \includegraphics[width=1\linewidth]{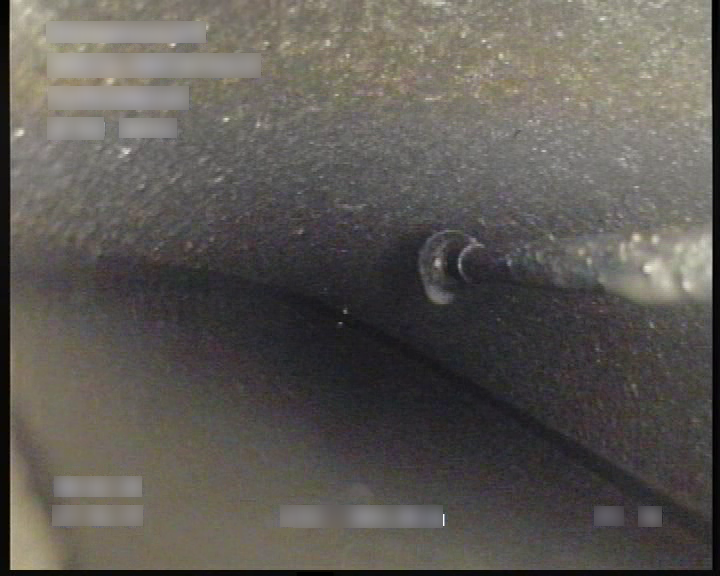} & \includegraphics[width=1\linewidth]{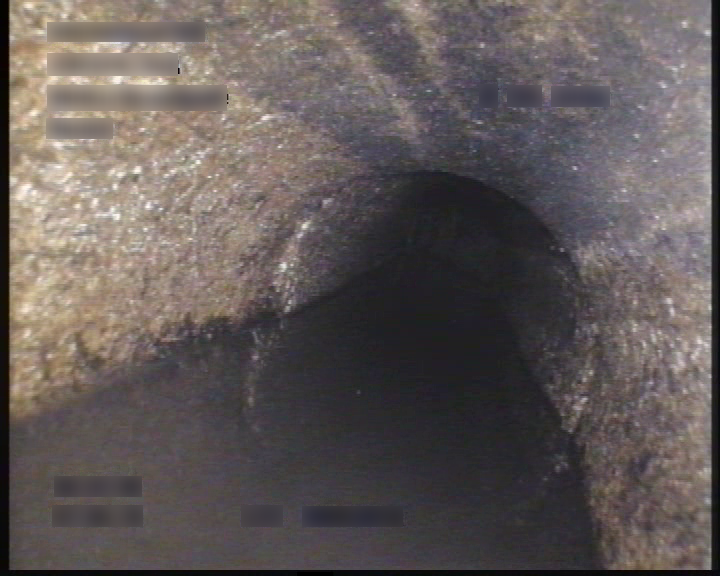} & \includegraphics[width=1\linewidth]{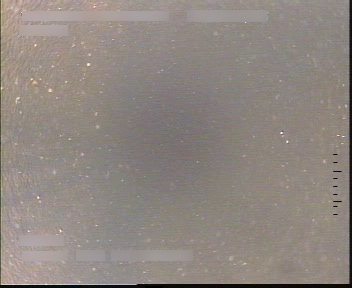}  \\
\label{tab:waterExamples}
\end{longtable}

\begin{longtable}[c]{m{1.5cm}m{2.69cm}m{2.69cm}m{2.69cm}m{2.69cm}m{2.69cm}}
\caption{\textbf{Pipe shape class examples.} Example images of the six considered pipe shape classes.}\\
\endfirsthead
\multicolumn{6}{c}%
        {{ Figure \thetable: \textbf{Continued from previous page}}} \\
\endhead
\multicolumn{6}{r}{{Continued on next page}} 
\endfoot
\endlastfoot
\centering
\textbf{Circular}     & \includegraphics[width=1\linewidth]{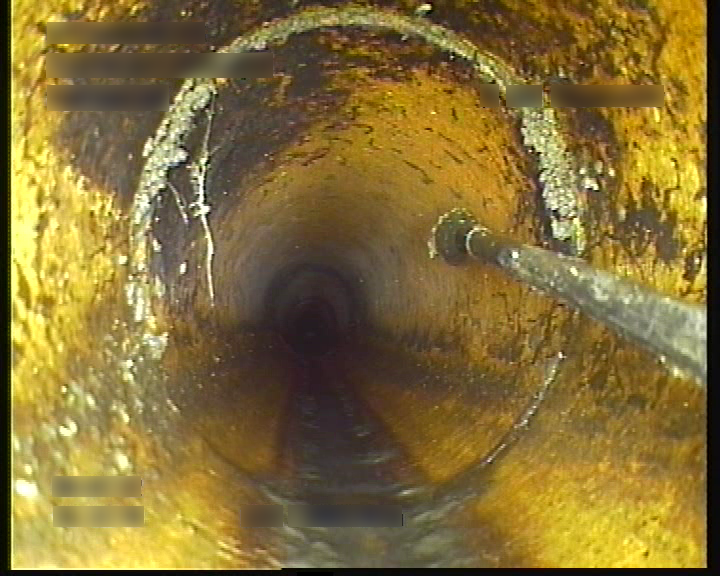} & \includegraphics[width=1\linewidth]{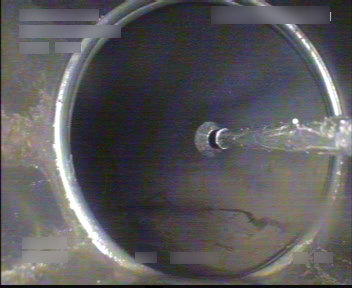} & \includegraphics[width=1\linewidth]{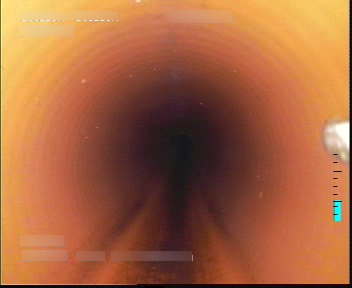} & \includegraphics[width=1\linewidth]{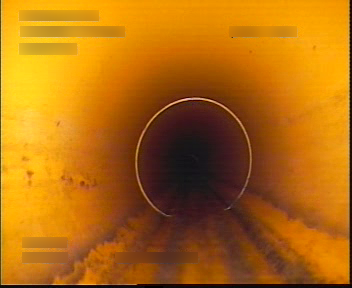} & \includegraphics[width=1\linewidth]{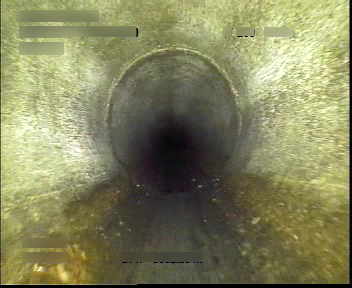}  \\
\textbf{Conical}     & \includegraphics[width=1\linewidth]{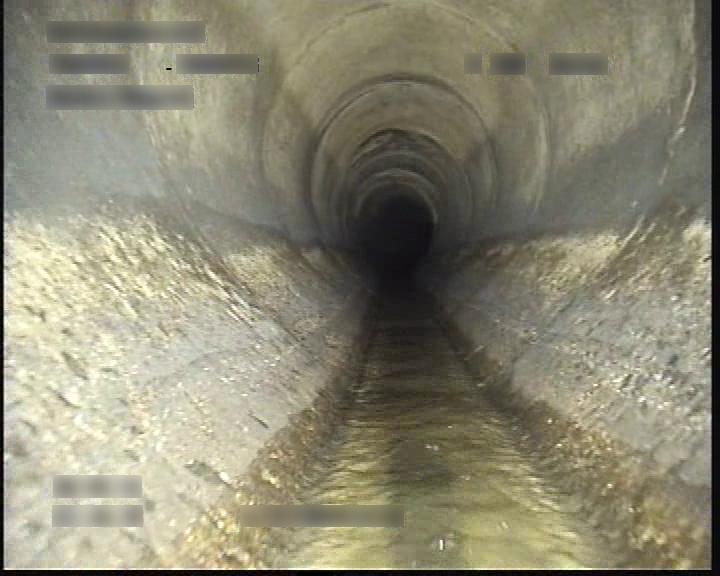} & \includegraphics[width=1\linewidth]{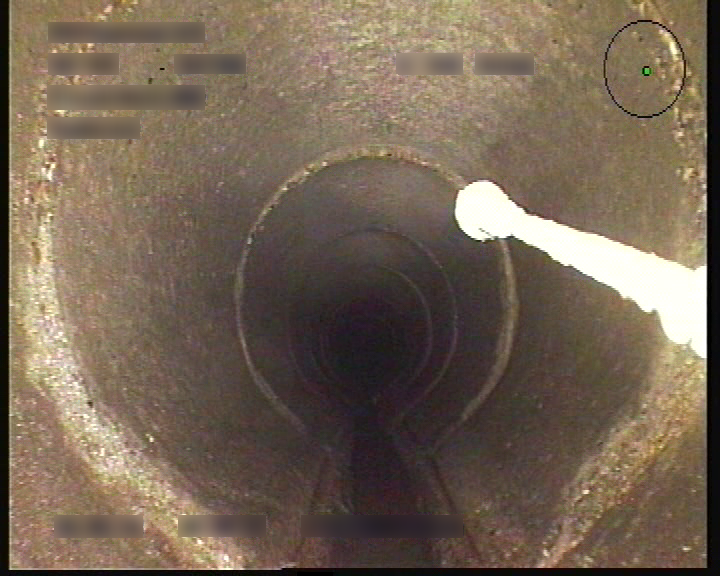} & \includegraphics[width=1\linewidth]{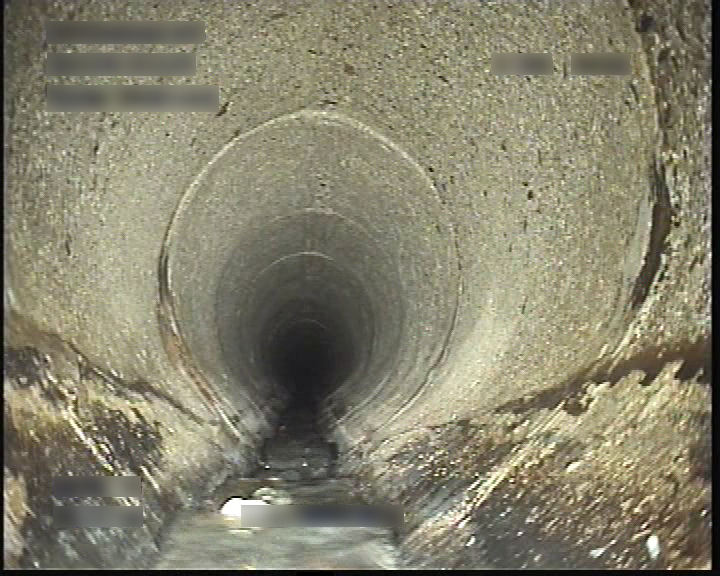} & \includegraphics[width=1\linewidth]{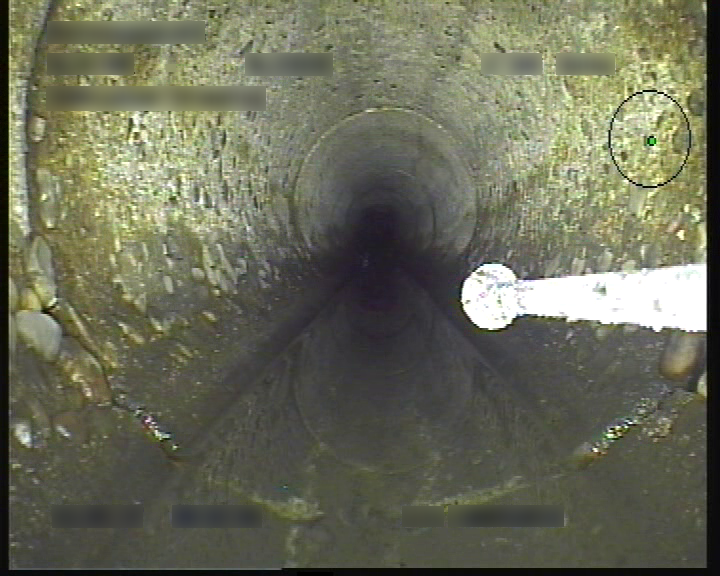} & \includegraphics[width=1\linewidth]{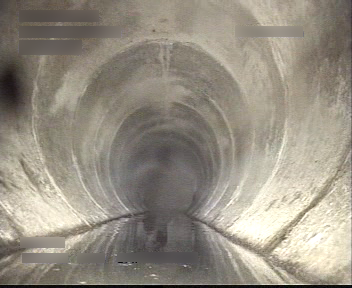}  \\

\textbf{Egg}     & \includegraphics[width=1\linewidth]{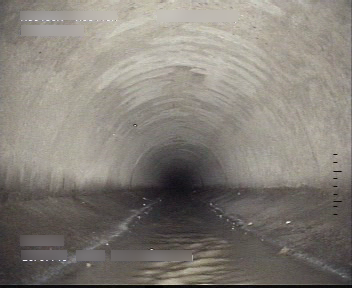} & \includegraphics[width=1\linewidth]{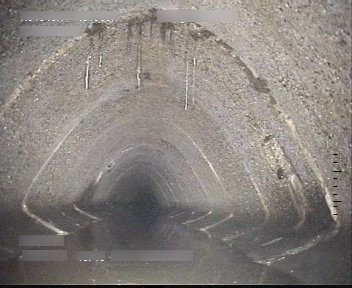} & \includegraphics[width=1\linewidth]{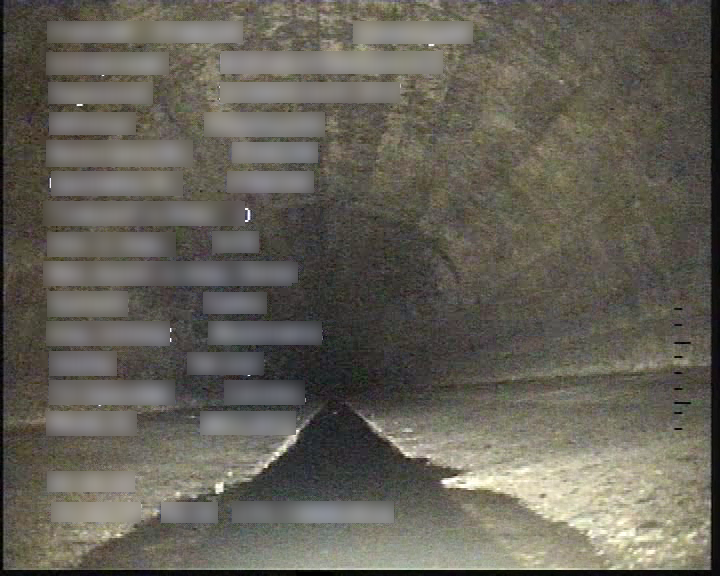} & \includegraphics[width=1\linewidth]{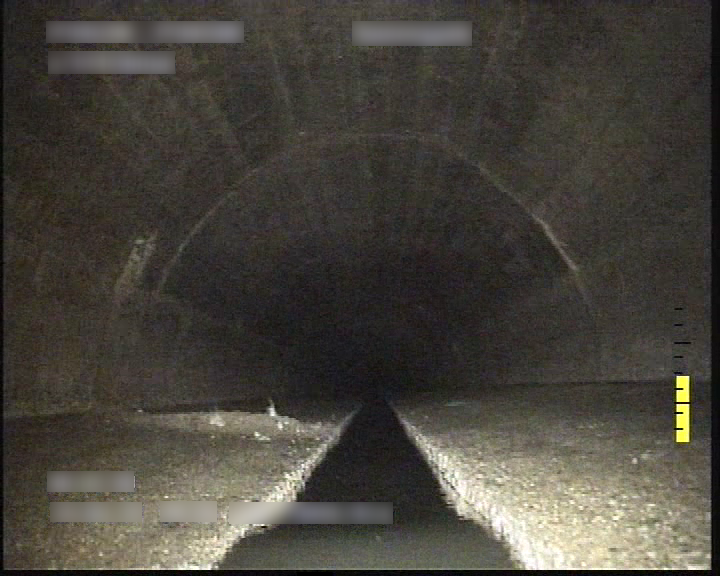} & \includegraphics[width=1\linewidth]{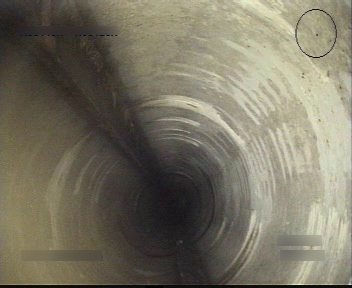}  \\

\textbf{Eye}     & \includegraphics[width=1\linewidth]{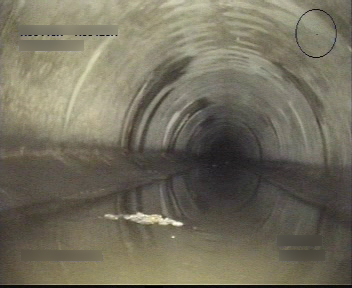} & \includegraphics[width=1\linewidth]{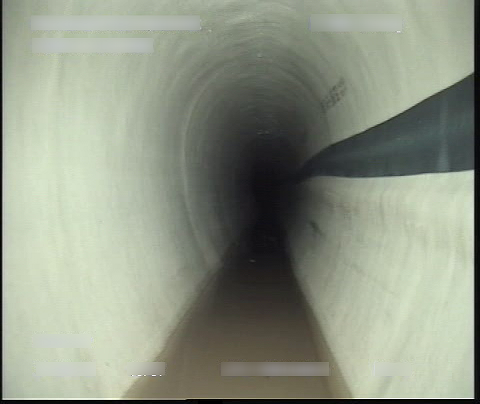} & \includegraphics[width=1\linewidth]{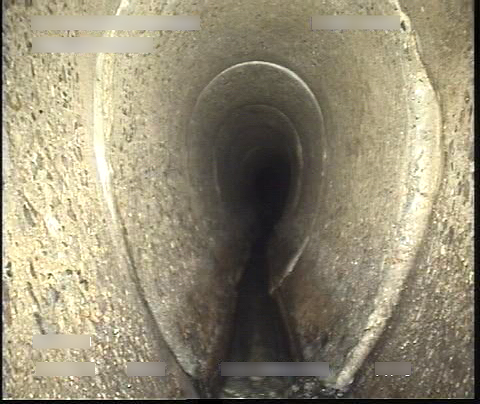} & \includegraphics[width=1\linewidth]{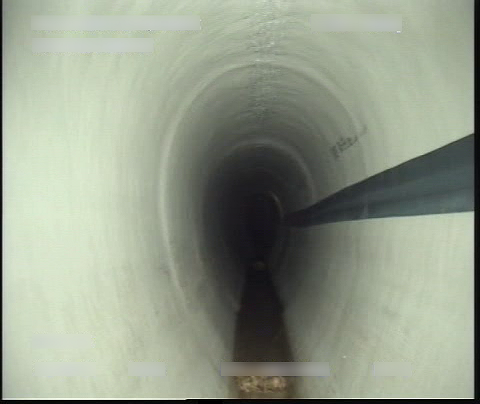} & \includegraphics[width=1\linewidth]{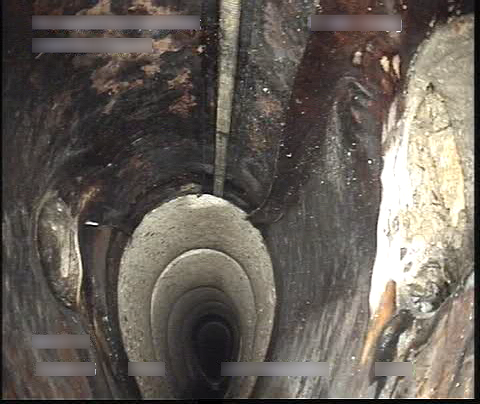}  \\
\textbf{Rectangular}     & \includegraphics[width=1\linewidth]{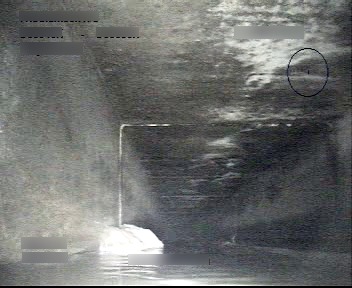} & \includegraphics[width=1\linewidth]{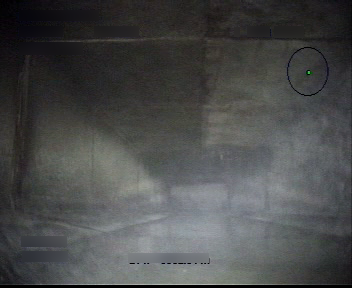} & \includegraphics[width=1\linewidth]{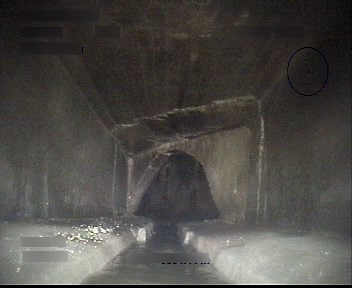} & \includegraphics[width=1\linewidth]{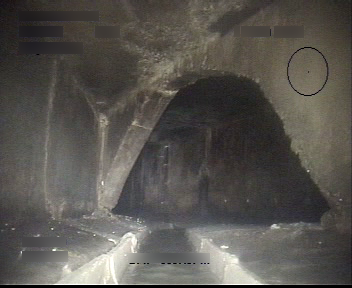} & \includegraphics[width=1\linewidth]{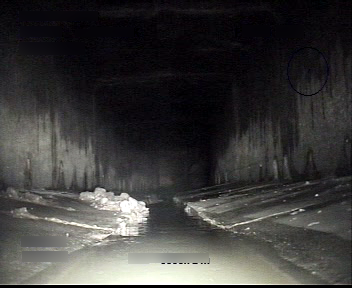}  \\
\textbf{Other}     &  
\includegraphics[width=1\linewidth]{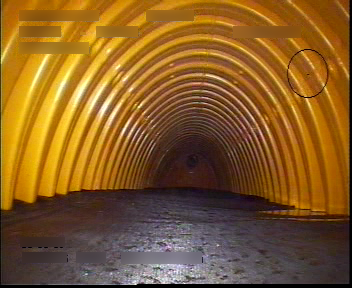} &
\includegraphics[width=1\linewidth]{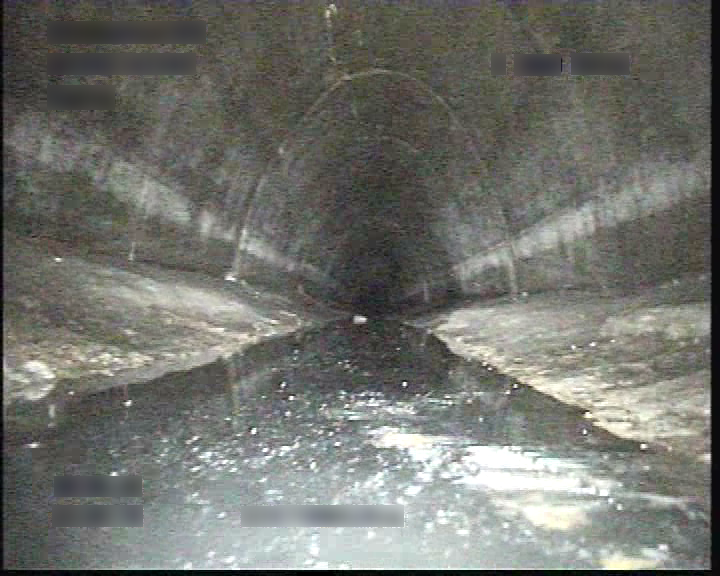} & \includegraphics[width=1\linewidth]{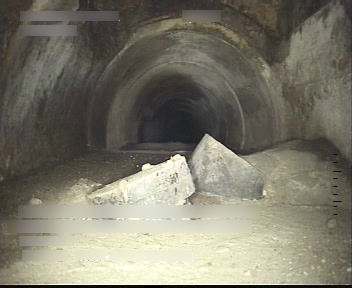} & \includegraphics[width=1\linewidth]{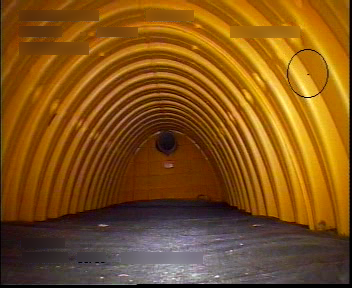} & \includegraphics[width=1\linewidth]{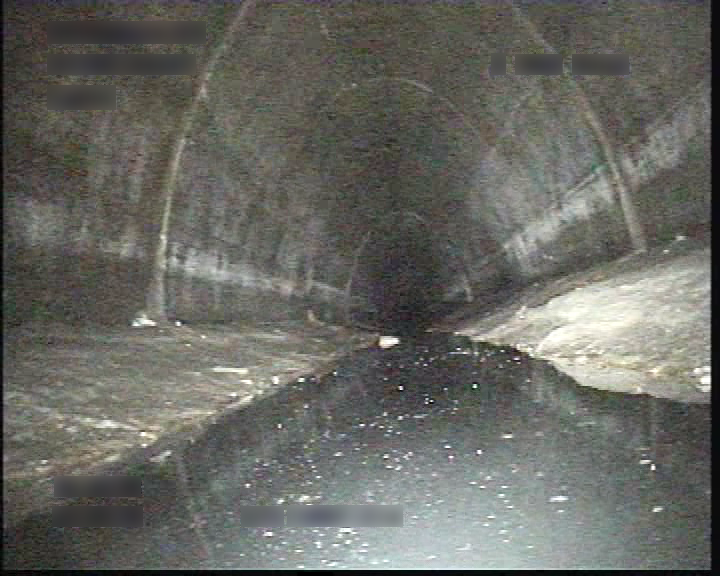}  \\
\label{tab:shapeExamples}
\end{longtable}

\begin{longtable}[c]{m{1.5cm}m{2.69cm}m{2.69cm}m{2.69cm}m{2.69cm}m{2.69cm}}
\caption{\textbf{Pipe material class examples.} Example images of the eight considered pipe material classes.}\\
\endfirsthead
\multicolumn{6}{c}%
        {{ Figure \thetable: \textbf{Continued from previous page}}} \\
\endhead
\multicolumn{6}{r}{{Continued on next page}} 
\endfoot
\endlastfoot
\centering

\textbf{Concrete}     & \includegraphics[width=1\linewidth]{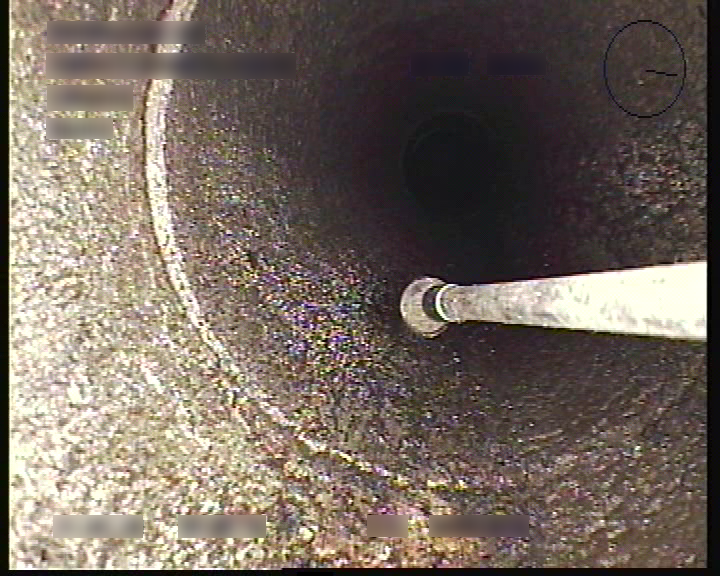} & \includegraphics[width=1\linewidth]{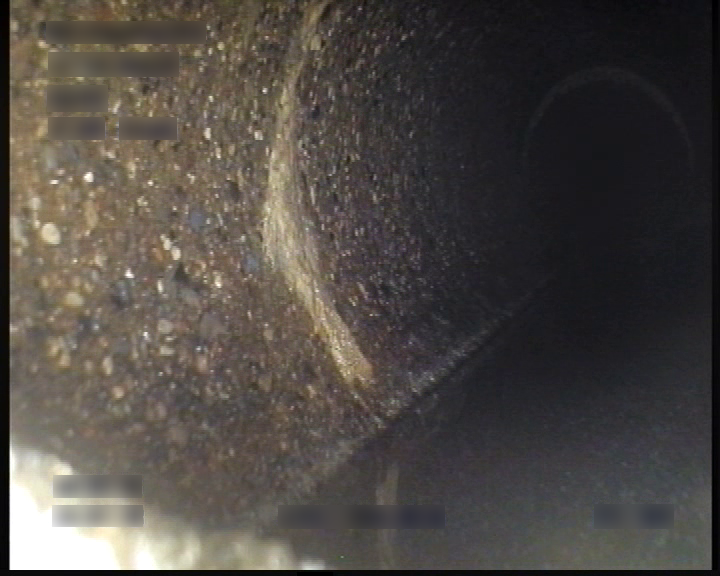} & \includegraphics[width=1\linewidth]{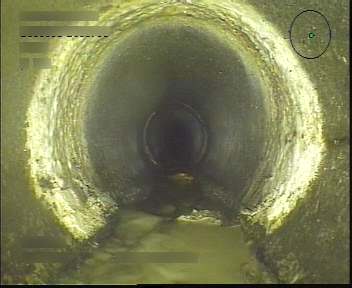} & \includegraphics[width=1\linewidth]{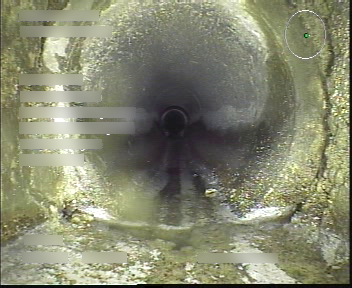} & \includegraphics[width=1\linewidth]{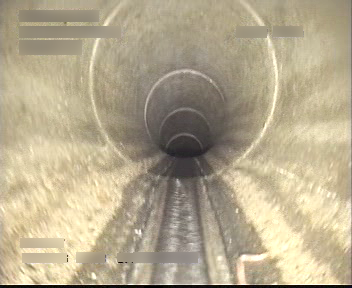}  \\

\textbf{Vitrified Clay}     &  
\includegraphics[width=1\linewidth]{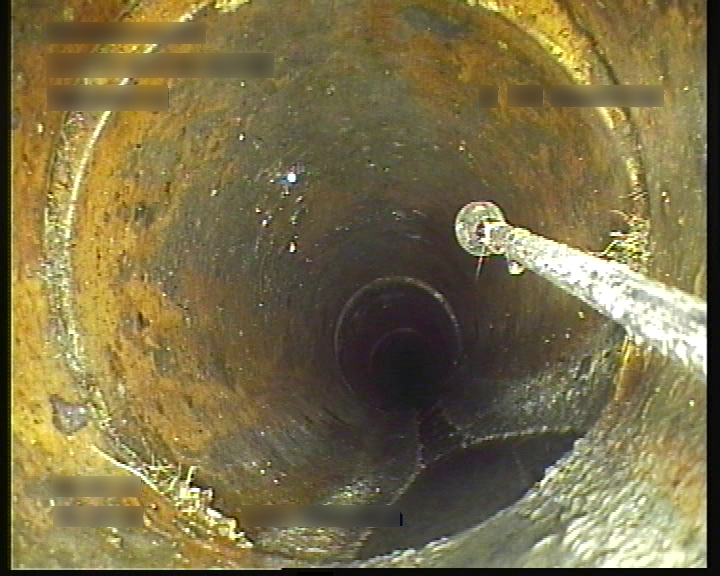} &
\includegraphics[width=1\linewidth]{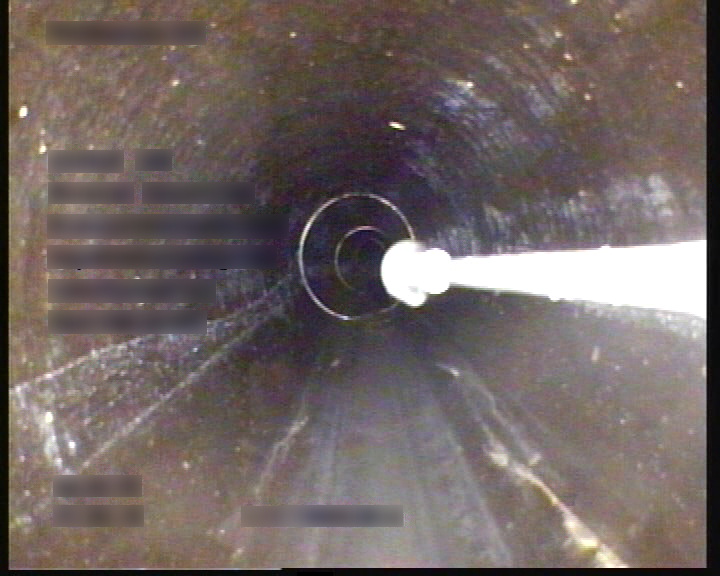} & \includegraphics[width=1\linewidth]{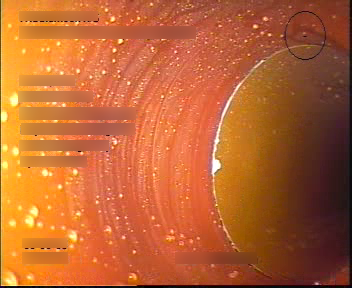} & \includegraphics[width=1\linewidth]{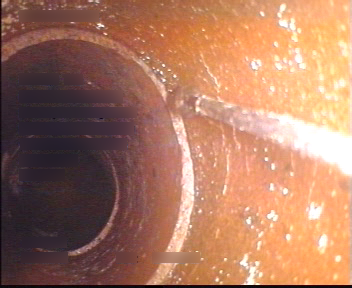} & \includegraphics[width=1\linewidth]{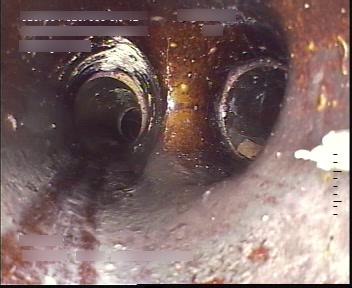}  \\

\textbf{Plastic}     & 
\includegraphics[width=1\linewidth]{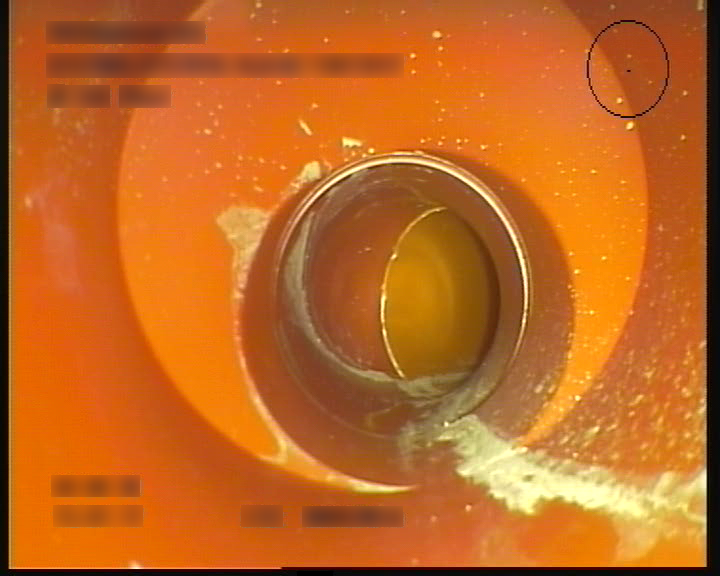} & \includegraphics[width=1\linewidth]{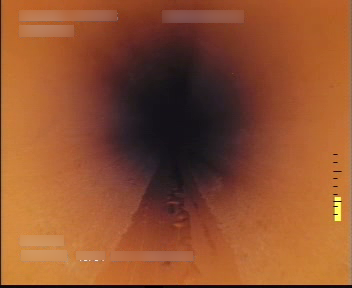} & \includegraphics[width=1\linewidth]{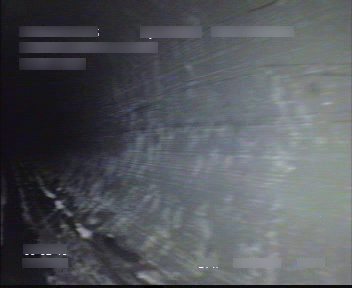} & \includegraphics[width=1\linewidth]{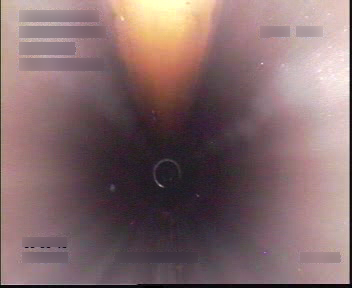} & \includegraphics[width=1\linewidth]{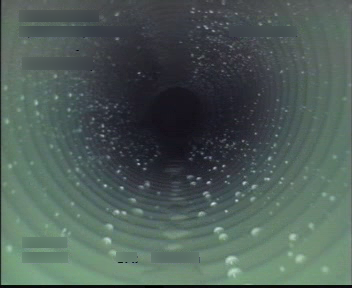}  \\

\textbf{Lining}     
& \includegraphics[width=1\linewidth]{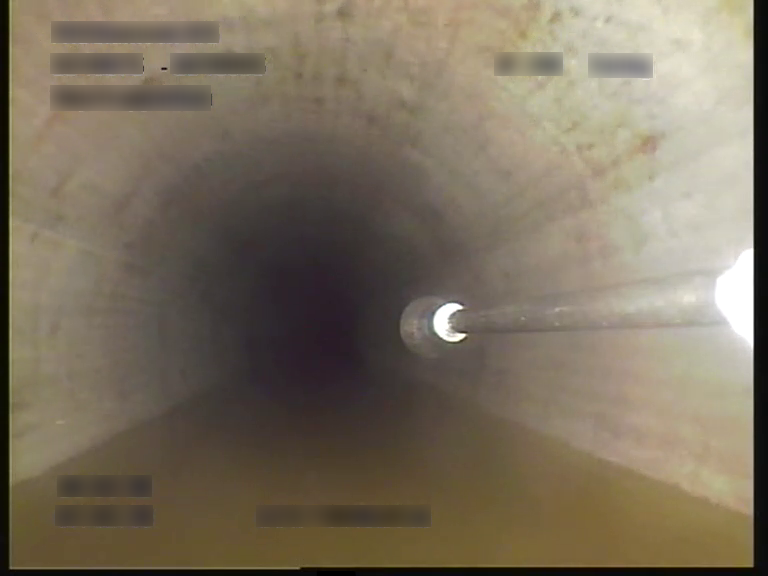} & \includegraphics[width=1\linewidth]{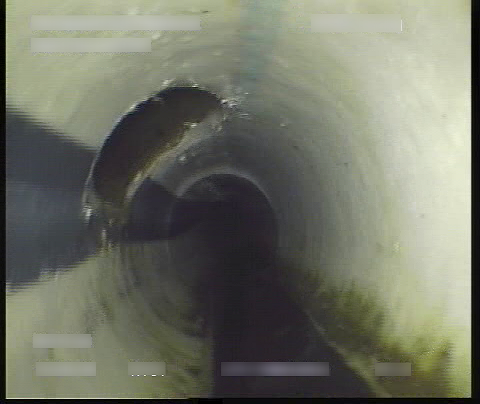} & \includegraphics[width=1\linewidth]{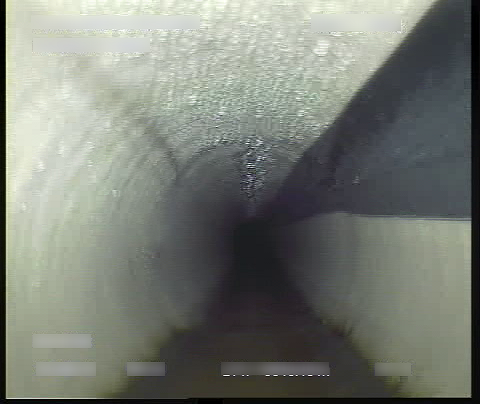} & \includegraphics[width=1\linewidth]{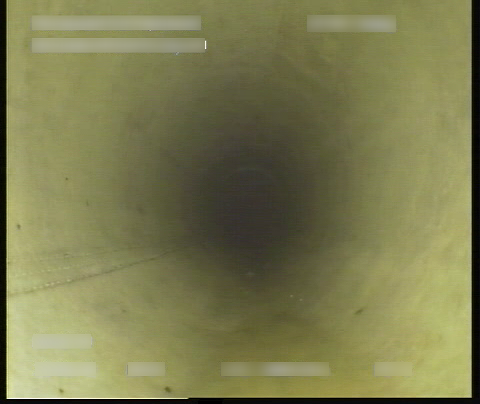} & \includegraphics[width=1\linewidth]{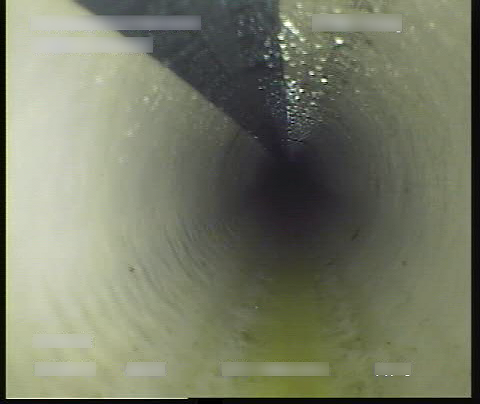}  \\

\textbf{Iron}     & \includegraphics[width=1\linewidth]{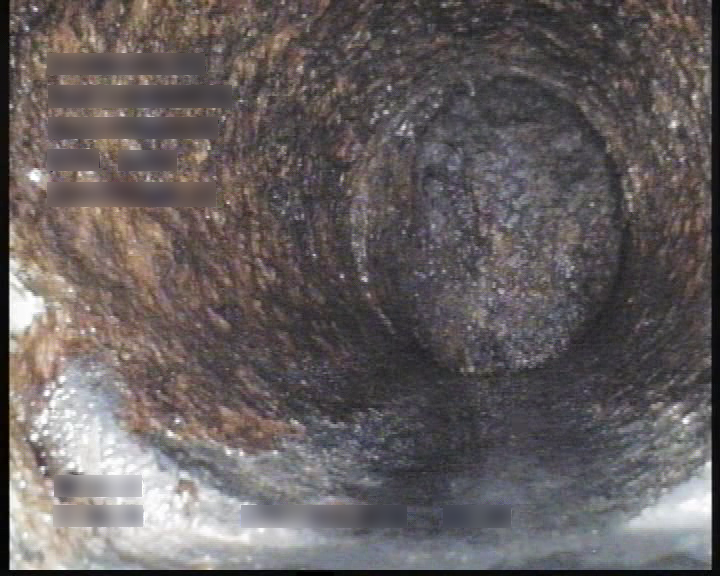} & \includegraphics[width=1\linewidth]{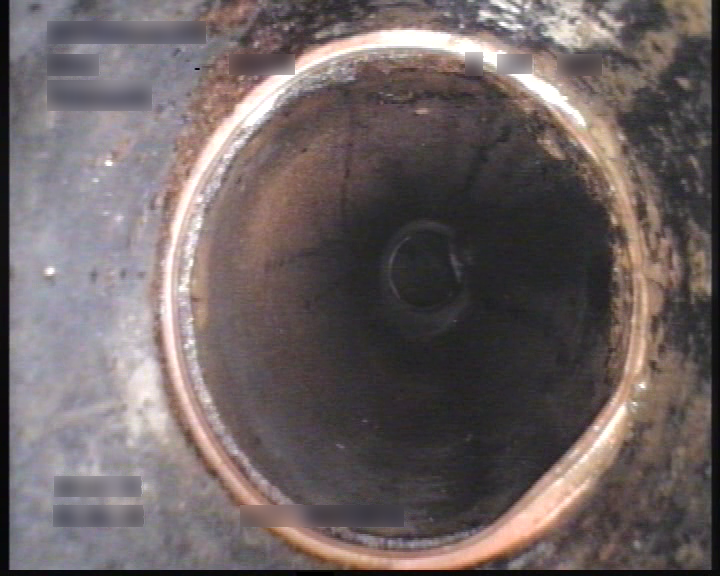} & \includegraphics[width=1\linewidth]{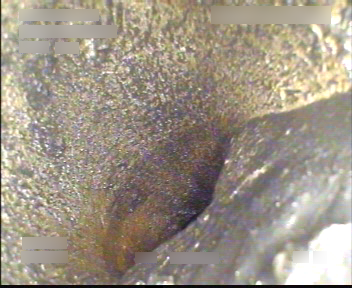} & \includegraphics[width=1\linewidth]{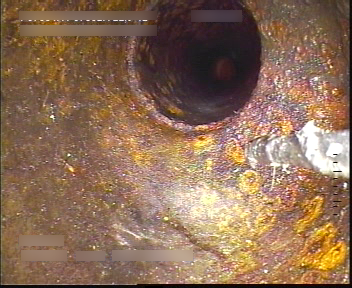} & \includegraphics[width=1\linewidth]{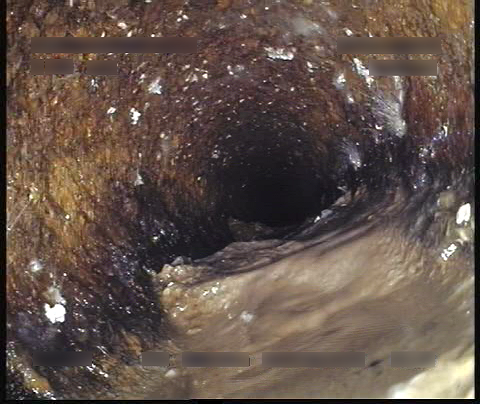}  \\

\textbf{Brickwork}     &  
\includegraphics[width=1\linewidth]{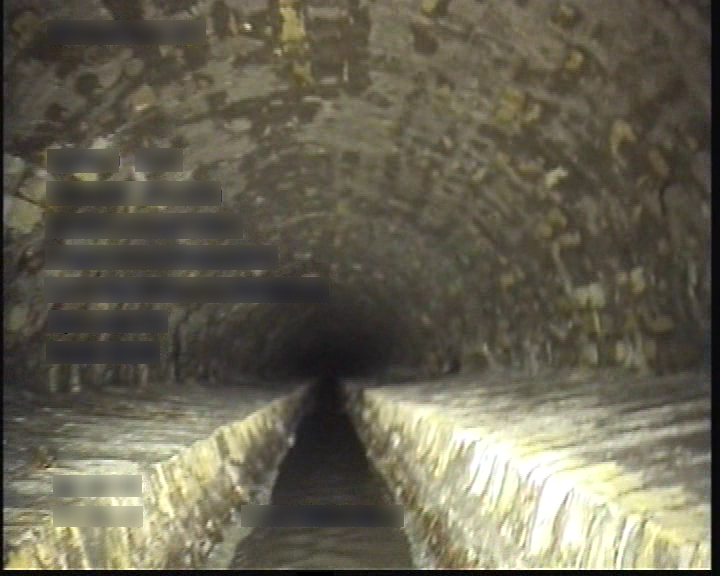} &
\includegraphics[width=1\linewidth]{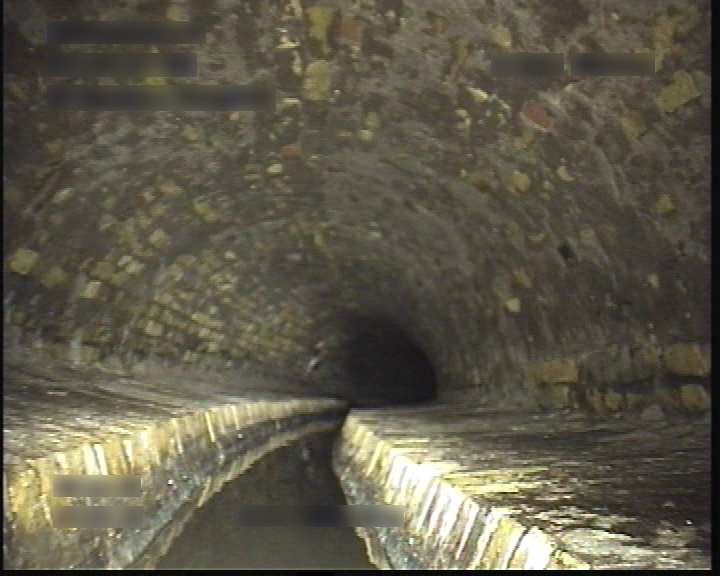} & \includegraphics[width=1\linewidth]{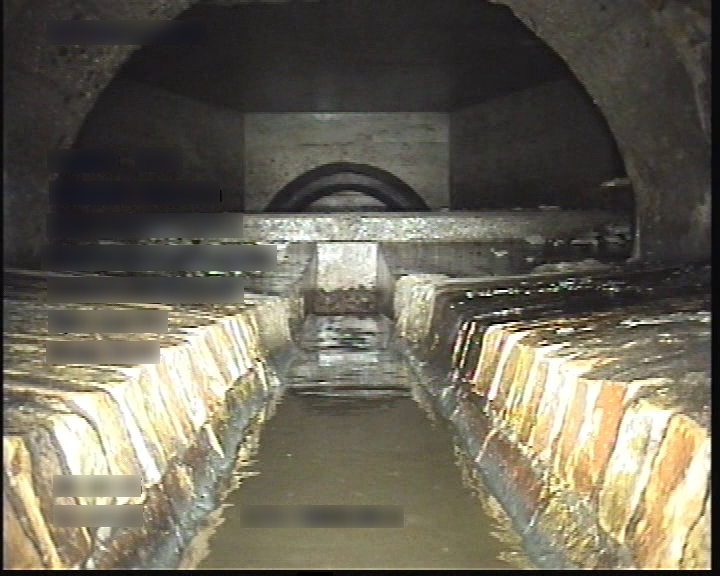} & \includegraphics[width=1\linewidth]{LaTeX/Figures/Shape_egg/00738911.png} & \includegraphics[width=1\linewidth]{LaTeX/Figures/Shape_egg/00738951.png}  \\

\textbf{Unknown}     & \includegraphics[width=1\linewidth]{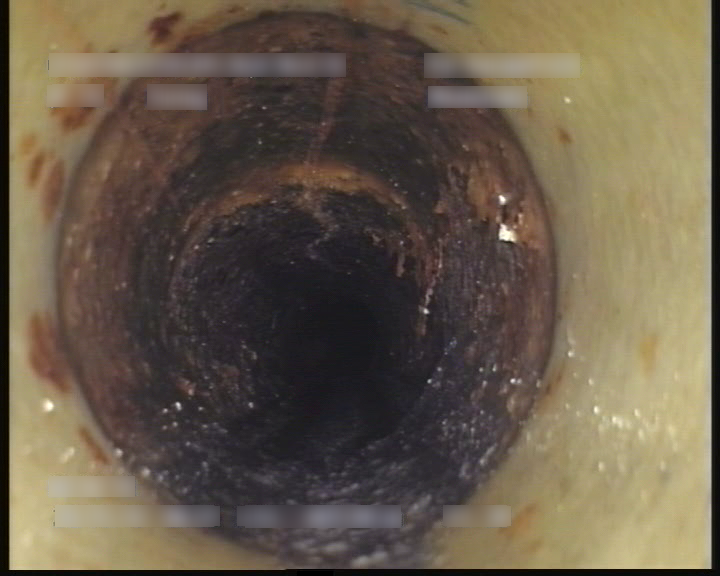} & \includegraphics[width=1\linewidth]{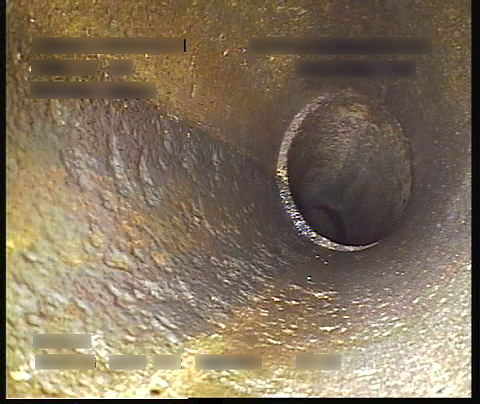} & \includegraphics[width=1\linewidth]{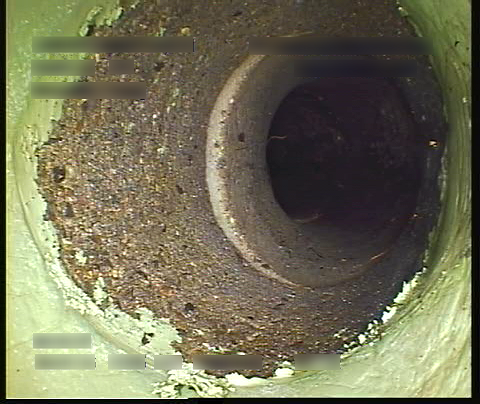} & \includegraphics[width=1\linewidth]{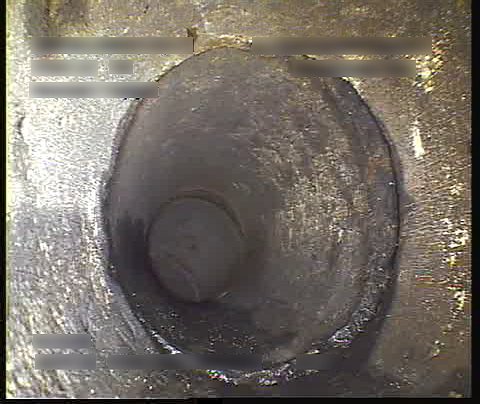} & \includegraphics[width=1\linewidth]{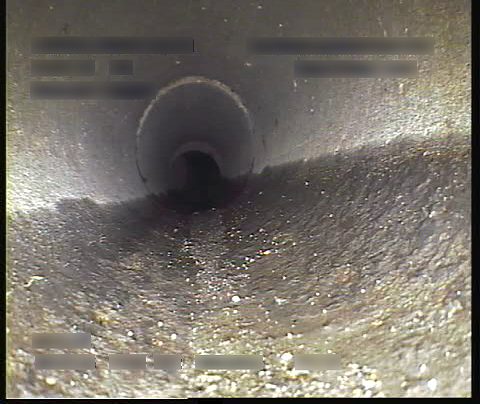}  \\

\textbf{Other}     & \includegraphics[width=1\linewidth]{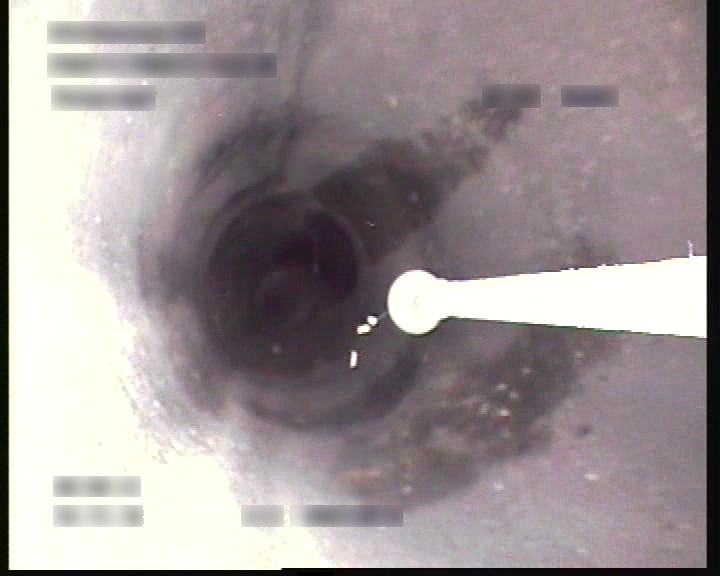} & \includegraphics[width=1\linewidth]{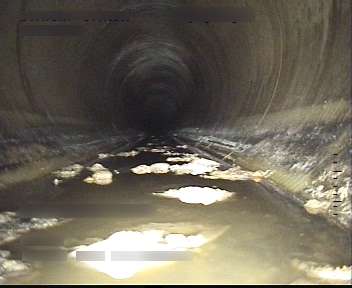} & \includegraphics[width=1\linewidth]{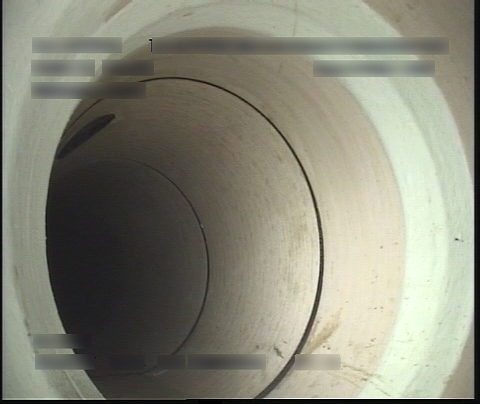} & \includegraphics[width=1\linewidth]{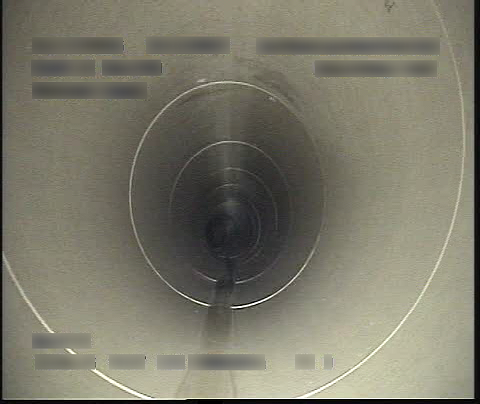} & \includegraphics[width=1\linewidth]{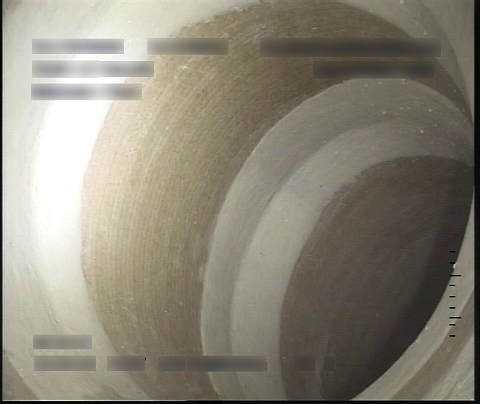}  \\

\label{tab:matExamples}
\end{longtable}

\twocolumn
\renewcommand\tablename{Table}

{\small
\bibliographystyle{ieee_fullname}
\bibliography{main}
}

\end{document}